\documentclass[11pt]{article}

\usepackage[preprint]{acl}

\usepackage{times}
\usepackage{latexsym}

\usepackage[T1]{fontenc}

\usepackage[utf8]{inputenc}

\usepackage{microtype}

\usepackage{inconsolata}

\usepackage{graphicx}

\usepackage{booktabs}
\usepackage{array} 

\definecolor{lightgreen}{RGB}{200,255,200}
\definecolor{lightyellow}{RGB}{255,255,200}
\definecolor{lightred}{RGB}{255,200,200}
\definecolor{lightblue}{RGB}{200,230,255}
\definecolor{lightpurple}{RGB}{230,200,255}
\definecolor{lightorange}{RGB}{255,230,200}
\usepackage{amsmath}
\usepackage{multirow} 
\usepackage{subfigure}

\usepackage{booktabs}
\usepackage{array}
\usepackage{multirow}
\usepackage{amsfonts,amssymb}
\usepackage{bm}
\usepackage{amsmath}
\usepackage{microtype}
\usepackage{bm}
\usepackage{pifont}
\usepackage{makecell}
\usepackage{graphicx}
\usepackage{booktabs}
\usepackage{subcaption}
\usepackage{caption}
\usepackage{float} 
\usepackage{adjustbox}
\usepackage{longtable}
\usepackage{amsmath}
\usepackage{fontawesome5}
\usepackage{color}
\usepackage{hyperref}

\title{Decoupling the Effect of Chain-of-Thought Reasoning:\\ A Human Label Variation Perspective}

\author{
 \textbf{Beiduo Chen\textsuperscript{\faMountain\kern1pt\faRobot}} \quad
 \textbf{Tiancheng Hu\textsuperscript{\faSchool}} \quad
 \textbf{Caiqi Zhang\textsuperscript{\faSchool}} \quad
 \textbf{Robert Litschko\textsuperscript{\faMountain\kern1pt\faRobot}} \\
 \textbf{Anna Korhonen\textsuperscript{\faSchool}} \quad
 \textbf{Barbara Plank\textsuperscript{\faMountain\kern1pt\faRobot}}
\\
\textsuperscript{\faMountain} MaiNLP, Center for Information and Language Processing, LMU Munich, Germany \\
\textsuperscript{\faRobot} Munich Center for Machine Learning (MCML), Munich, Germany \\
\textsuperscript{\faSchool} Language Technology Lab, University of Cambridge, United Kingdom \\
{\tt \{\href{mailto:beiduo.chen@lmu.de}{\textcolor{black}{beiduo.chen}}, \href{mailto:robert.litschko@lmu.de}{\textcolor{black}{robert.litschko}}, \href{mailto:b.plank@lmu.de}{\textcolor{black}{b.plank}}\}@lmu.de}
\\
{\tt \{\href{mailto:th656@cam.ac.uk}{\textcolor{black}{th656}}, \href{mailto:cz391@cam.ac.uk}{\textcolor{black}{cz391}}, \href{mailto:alk23@cam.ac.uk}{\textcolor{black}{alk23}}\}@cam.ac.uk}
}

\begin{document}
\maketitle

\begin{abstract} 

Reasoning-tuned LLMs utilizing long Chain-of-Thought (CoT) excel at single-answer tasks, yet their ability to model Human Label Variation—which requires capturing probabilistic ambiguity rather than resolving it—remains underexplored. We investigate this through systematic disentanglement experiments on distribution-based tasks, employing Cross-CoT experiments to isolate the effect of reasoning text from intrinsic model priors. We observe a distinct "decoupled mechanism": while CoT improves distributional alignment, final accuracy is dictated by CoT content (99\% variance contribution), whereas distributional ranking is governed by model priors (over 80\%). Step-wise analysis further shows that while CoT’s influence on accuracy grows monotonically during the reasoning process, distributional structure is largely determined by LLM’s intrinsic priors. These findings suggest that long CoT serves as a decisive LLM decision-maker for the top option but fails to function as a granular distribution calibrator for ambiguous tasks. 

\end{abstract}

\section{Introduction}

Reasoning-tuned large language models (LLMs) with long CoT reasoning achieve strong performance on many benchmarks~\cite{DBLP:journals/corr/abs-2307-09288,DBLP:journals/corr/abs-2407-21783,DBLP:journals/corr/abs-2303-08774,DBLP:conf/nips/Wei0SBIXCLZ22,DBLP:conf/iclr/0002WSLCNCZ23,DBLP:journals/corr/abs-2501-12948,qwq32b,DBLP:journals/corr/abs-2410-21276}, usually measured by accuracy under the assumption of a single correct answer~\cite{DBLP:conf/iclr/HendrycksBBZMSS21,DBLP:journals/corr/abs-2311-12022,DBLP:conf/nips/WangMZNCGRAHJLK24,DBLP:journals/corr/abs-2503-21380,DBLP:conf/nips/HendrycksBKABTS21}. However, many real-world tasks are inherently ambiguous or subjective, with human annotators often disagreeing due to genuine semantic uncertainty~\cite{pavlick-kwiatkowski-2019-inherent,DBLP:journals/aim/AroyoW15}. Such Human Label Variation (HLV) requires models to predict distributions over plausible answers, making argmax-based evaluation insufficient~\cite{DBLP:journals/jair/UmaFHPPP21,plank-2022-problem,cabitza2023toward, DBLP:journals/corr/abs-2510-17516}.
Intuitively, reasoning through intermediate steps might better reflect such variations compared to direct answering~\cite{chen-etal-2025-threading}, motivating us to ask \textit{RQ1: whether long CoT helps models better approximate human label distributions}, and \textit{RQ2: whether any gains come from CoT reasoning or the model’s latent parametric knowledge.}

To investigate \textit{RQ1}, we utilize ChaosNLI~\cite{nie-etal-2020-learn}, a benchmark capturing collective human opinions. We analyze the latent answer distributions behind CoT using complementary metrics: accuracy for correctness, and Jensen–Shannon Divergence (JSD, \citealt{DBLP:journals/tit/EndresS03}) and Spearman’s $\rho$ \cite{spearman1961proof} for distributional and ranking alignment. To further disentangle CoT’s role from model-intrinsic priors (\textit{RQ2}), we conduct: $\mathrm{i}$) {Cross-CoT experiments}, injecting one model’s CoT into another to test reasoning transfer; and $\mathrm{ii}$) {Step-wise analysis}, truncating CoT to track how influence evolves over reasoning steps.

Our analysis uncovers a notable ``split influence''. While LLMs generally improve distributional alignment (lower JSD) after reasoning, this gain is not uniform across metrics. Using ANOVA to calculate the variance contribution percentage in our Cross-CoT experiments, we find that final accuracy is overwhelmingly determined by the CoT content ($\approx$99\%), confirming the strong role of reasoning chain in steering the top-1 answer decision.
In stark contrast, the distributional structure—ranking and probability allocation among non-argmax options—is largely immune to CoT, remaining governed by model priors (>80\%). 

Step-wise analysis further clarifies this dynamic. While all metrics evolve throughout reasoning, changes in accuracy are predominantly driven by CoT and grow monotonically with later steps. By comparison, changes in distributional similarity (JSD and Spearman’s $\rho$) are mostly determined by the LLM’s intrinsic behavior. 
This reveals a dichotomy: current long CoT paradigms act as strong LLM decision makers but weak distribution calibrators. 
CoT tends to progressively concentrate probability mass to lock in the most likely answer latently, but fails to govern the reshaping of the probability landscape for alternative options. 
This work highlights the structural limitations of current reasoning processes in capturing fine-grained answer uncertainty and motivates the need for distribution-aware reasoning mechanisms.

\section{Background}

\paragraph{HLV in Natural Language Inference.}
Unlike the single-label assumption in most benchmarks, NLI is often inherently ambiguous: a premise and hypothesis can elicit a spectrum of plausible interpretations, a phenomenon known as HLV~\cite{plank-2022-problem}. Benchmarks such as ChaosNLI capture this by representing labels as probability distributions rather than single gold labels~\cite{nie-etal-2020-learn,weber-genzel-etal-2024-varierr,jiang-etal-2023-ecologically,hong-etal-2025-litex,DBLP:journals/corr/abs-2510-16458}. Evaluating models under HLV requires moving beyond standard accuracy to distributional metrics that measure alignment with collective human judgments~\cite{DBLP:journals/corr/abs-2502-01891,lee-etal-2023-large,leonardelli-etal-2023-semeval,chen-etal-2024-seeing,chen-etal-2025-rose,chen-etal-2025-threading,ni2025can}.

\paragraph{Reasoning under Distributional Uncertainty.}
Recent LLM advancements emphasize reasoning-intensive paradigms. Long CoT enables models to decompose problems into intermediate steps~\cite{DBLP:conf/iclr/0002WSLCNCZ23,DBLP:journals/corr/abs-2501-12948,qwq32b,DBLP:journals/corr/abs-2410-21276}, effectively reducing uncertainty and producing high-confidence conclusions in deterministic tasks. However, its role in probabilistic HLV settings is less clear. Generating explicit reasoning can inadvertently suppress valid alternative interpretations, potentially biasing the model toward the top-1 choice. While prior work has explored confidence-based calibration~\cite{DBLP:journals/corr/abs-2508-01191,DBLP:journals/corr/abs-2505-14489,DBLP:journals/corr/abs-2506-08243}, it remains unclear whether CoT actively shapes the full output distribution or mainly rationalizes the final decision, leaving non-argmax probabilities governed by the model’s intrinsic priors.

\section{Experiments}

\subsection{Setup}
\paragraph{Task}
We experiment on 3 ChaosNLI subsets: MNLI, SNLI, and $\alpha$NLI~\cite{bowman-etal-2015-large,williams-etal-2018-broad,DBLP:conf/iclr/BhagavatulaBMSH20}. Each instance is annotated by 100 crowdworkers, enabling reliable human judgment distributions (HJD). MNLI and SNLI are three-way classification tasks (entailment, neutral, contradiction), yielding 3-d label distributions. $\alpha$NLI is a binary-choice task, where annotators select the better hypothesis for a given observation pair, producing 2-d distributions.\!\footnote{ChaosNLI is ideal for HLV evaluation as a rare non–social science benchmark with collective HJDs~\cite{DBLP:journals/corr/abs-2510-17516}.} Dataset details are in Appendix~\ref{app:dataset}.

\begin{table}[t]
\centering
\resizebox{\linewidth}{!}{
\begin{tabular}{ll}
\toprule
\textbf{Reasoning  LLMs}                           & \textbf{Abbr.}    \\
\midrule
Qwen/Qwen3-30B-A3B-Thinking-2507~\cite{qwen3technicalreport}          & Qwen     \\
deepseek-ai/DeepSeek-R1-Distill-Llama-70B~\cite{DBLP:journals/corr/abs-2501-12948} & R1-Llama \\
deepseek-ai/DeepSeek-R1-Distill-Qwen-32B~\cite{DBLP:journals/corr/abs-2501-12948}  & R1-Qwen  \\
allenai/Olmo-3-32B-Think~\cite{olmo2025olmo}                  & Olmo     \\
zai-org/GLM-Z1-32B-0414~\cite{glm2024chatglm}                   & GLM      \\
ByteDance-Seed/Seed-OSS-36B-Instruct~\cite{seed2025seed-oss}      & Seed     \\
openai/gpt-oss-20b~\cite{openai2025gptoss120bgptoss20bmodel}                        & GPT     \\
\bottomrule
\end{tabular}}
\caption{Reasoning LLMs and their abbreviation.}
\label{tab:llms}
\end{table}

\paragraph{Models}
To comprehensively evaluate the HLV performance of reasoning-tuned LLMs, we select a range of state-of-the-art open-source reasoning models (details in Table~\ref{tab:llms}).
All follow a reason-then-answer paradigm: generating a long CoT reasoning process before outputting a final answer.

\paragraph{Evaluation}
All NLI instances are reformulated as multiple-choice questions. Model predictions are extracted using the first-token probability method~\cite{DBLP:conf/icml/SanturkarDLLLH23,DBLP:journals/corr/abs-2306-16388, DBLP:journals/tmlr/LiangBLTSYZNWKN23}, where logits are aggregated and normalized to obtain an output probability distribution over answer options.\!\footnote{While this approximation may not fully capture downstream decoding 
dynamics, our analysis focuses on relative comparisons across controlled conditions, where this proxy is consistently applied.}
We measure the HLV alignment between the model-generated distribution and the corresponding HJD using JSD. We also report accuracy.
Additionally, we include Spearman’s $\rho$ in Section~\ref{sec:Last_CoT}, which is invariant to monotonic transformations.
See details in Appendix~\ref{app:evaluation}.

\newcommand{\hlc}[2]{\setlength{\fboxsep}{1pt}\colorbox{#1}{#2}}

\begin{table*}[t]
\centering
\resizebox{\linewidth}{!}{
\begin{tabular}{lcccccccccccc}
\toprule
\textbf{Task}        & \multicolumn{4}{c}{MNLI}                        & \multicolumn{4}{c}{SNLI}                        & \multicolumn{4}{c}{${\alpha}$NLI}                    \\ \cmidrule(lr){1-1} \cmidrule(lr){2-5} \cmidrule(lr){6-9} \cmidrule(lr){10-13} 
\textbf{LLMs/Metrics} & ACC$_{start}$ ↑ & ACC$_{last}$ ↑& JSD$_{start}$ ↓ & JSD$_{last}$ ↓& ACC$_{start}$ ↑ & ACC$_{last}$ ↑& JSD$_{start}$ ↓ & JSD$_{last}$ ↓& ACC$_{start}$ ↑ & ACC$_{last}$ ↑& JSD$_{start}$ ↓ & JSD$_{last}$ ↓ \\ \midrule
Qwen        & 0,688      & 0,644     & 0,093      & \hlc{lightblue}{0,080}      & 0,668      & \hlc{lightred}{0,778}     & 0,144      & \hlc{lightblue}{0,119}     & 0,749      & \hlc{lightred}{0,890}      & 0,108      & \hlc{lightblue}{0,084}     \\
R1-Llama    & 0,666      & \hlc{lightred}{0,689}     & 0,082      & \hlc{lightblue}{0,077}     & 0,615      & \hlc{lightred}{0,750}      & 0,133      & \hlc{lightblue}{0,123}     & 0,839      & \hlc{lightred}{0,878}     & 0,098      & \hlc{lightblue}{0,091}     \\
R1-Qwen     & 0,734      & 0,672     & 0,080       & \hlc{lightblue}{0,072}     & 0,689      & \hlc{lightred}{0,764}     & 0,127      & \hlc{lightblue}{0,115}     & 0,832      & \hlc{lightred}{0,860}      & 0,094      & \hlc{lightblue}{0,081}     \\
Olmo        & 0,614      & 0,609     & 0,088      & \hlc{lightblue}{0,082}     & 0,738      & \hlc{lightred}{0,775}     & 0,133      & \hlc{lightblue}{0,122}     & 0,819      & \hlc{lightred}{0,863}     & 0,107      & \hlc{lightblue}{0,087}     \\
GLM         & 0,670       & 0,640      & 0,082      & \hlc{lightblue}{0,077}     & 0,545      & \hlc{lightred}{0,756}     & 0,134      & \hlc{lightblue}{0,120}      & 0,834      & \hlc{lightred}{0,888}     & 0,099      & \hlc{lightblue}{0,088}     \\
Seed        & 0,705      & 0,614     & 0,077      & 0,083     & 0,766      & \hlc{lightred}{0,777}     & 0,124      & 0,127     & 0,868      & \hlc{lightred}{0,887}     & 0,098      & \hlc{lightblue}{0,095}     \\
GPT         & 0,437      & \hlc{lightred}{0,672}     & 0,095      & \hlc{lightblue}{0,077}     & 0,596      & \hlc{lightred}{0,772}     & 0,145      & \hlc{lightblue}{0,119}     & 0,793      & \hlc{lightred}{0,872}     & 0,112      & \hlc{lightblue}{0,080}     \\
\bottomrule
\end{tabular}}
\caption{Results before and after reasoning. \textit{start} and \textit{last} denote before reasoning and after completion. Red indicates an increase, blue a decrease. Arrows next to metric names show whether higher or lower is better.}
\label{tab:Q1}
\end{table*}

\subsection{Does CoT Improve HLV Performance?}

We examine the impact of reasoning by comparing model performance before and after CoT. See Table~\ref{tab:Q1}, the effect of CoT on accuracy is mixed. While most models improve on SNLI and $\alpha$NLI, performance on MNLI is highly unstable. In contrast, JSD consistently decreases across nearly all models and datasets. Importantly, this improved distributional alignment is often decoupled from accuracy: even when the accuracy decreases (e.g., Qwen on MNLI), the output distribution aligns more closely with human judgments on average.

CoT generally reduces JSD, indicating useful signals for HLV, but the benefit varies across models. Models with similar post-CoT accuracy can exhibit substantially different JSD values, and vice versa, raising the key attribution question: \textit{are the gains driven by the semantic content of the reasoning itself, or by model-specific inductive biases when interpreting the reasoning text?}

\begin{figure}[t]
    \centering
    \subfigure[Delta Accuracy ↑]{
        \includegraphics[width=0.9\linewidth]{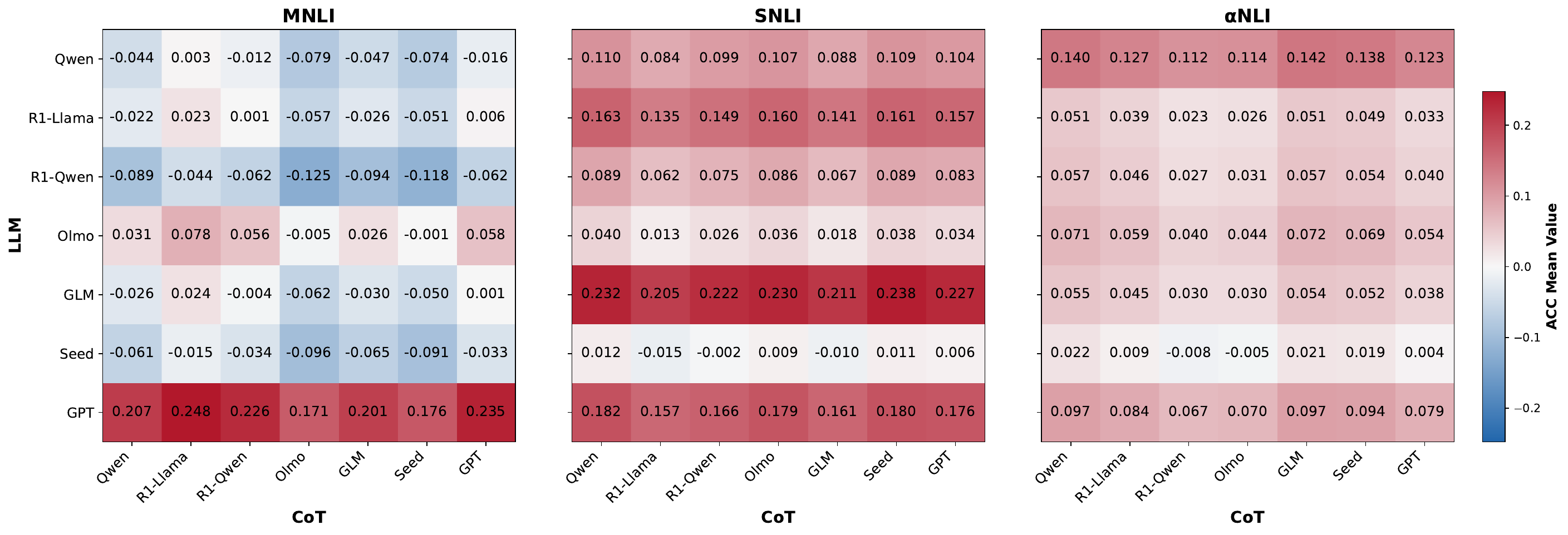}\label{fig:delta-acc}
    }
    
    \subfigure[Delta JSD ↓]{
        \includegraphics[width=0.9\linewidth]{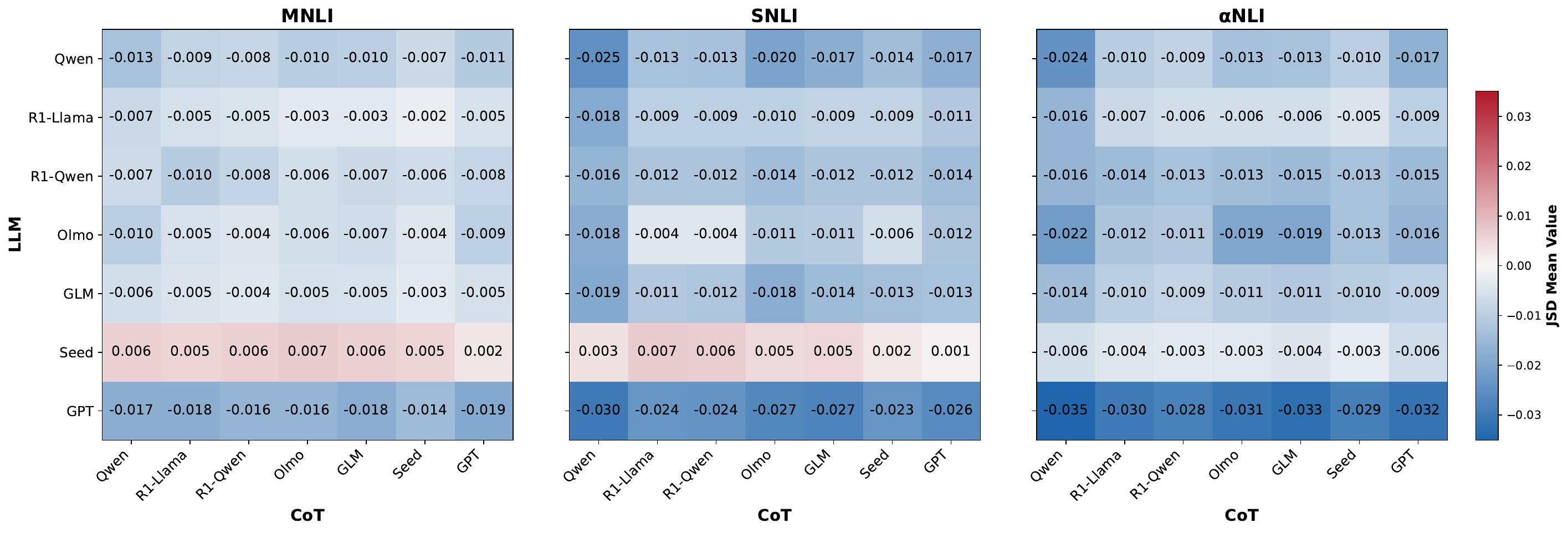}\label{fig:delta-jsd}
    }
    \caption{Changes after reasoning for Cross-CoT.}
    \label{fig:delta}
\end{figure}

\subsection{Cross-CoT Evaluation}

To disentangle the source of JSD improvements, we conduct \textit{Cross-CoT} experiments, injecting reasoning paths from different source models into various inference models. 
We show the performance changes in Figure~\ref{fig:delta}. On MNLI, accuracy shows mixed patterns. In contrast, consistent with single-model results, JSD improves across nearly all Cross-CoT pairings.\!\footnote{The box plot (Figure~\ref{fig:delta-jsd-box}) in Appendix~\ref{app:Cross-CoT} shows improvements are widespread across instances, not driven by outliers.} Regardless of the reasoning source model, injecting a CoT almost universally reduces divergence from human distributions. This confirms: \textbf{CoT text acts as a portable carrier of HLV-relevant information—reasoning generated by one model can facilitate better distributional alignment in another.}
However, the divergent patterns between accuracy and JSD motivate us to examine why these two metrics respond differently to CoT, and how CoT influences them.

\begin{figure}[t]
    \centering
    \subfigure[Last Accuracy ↑]{
        \includegraphics[width=0.95\linewidth]{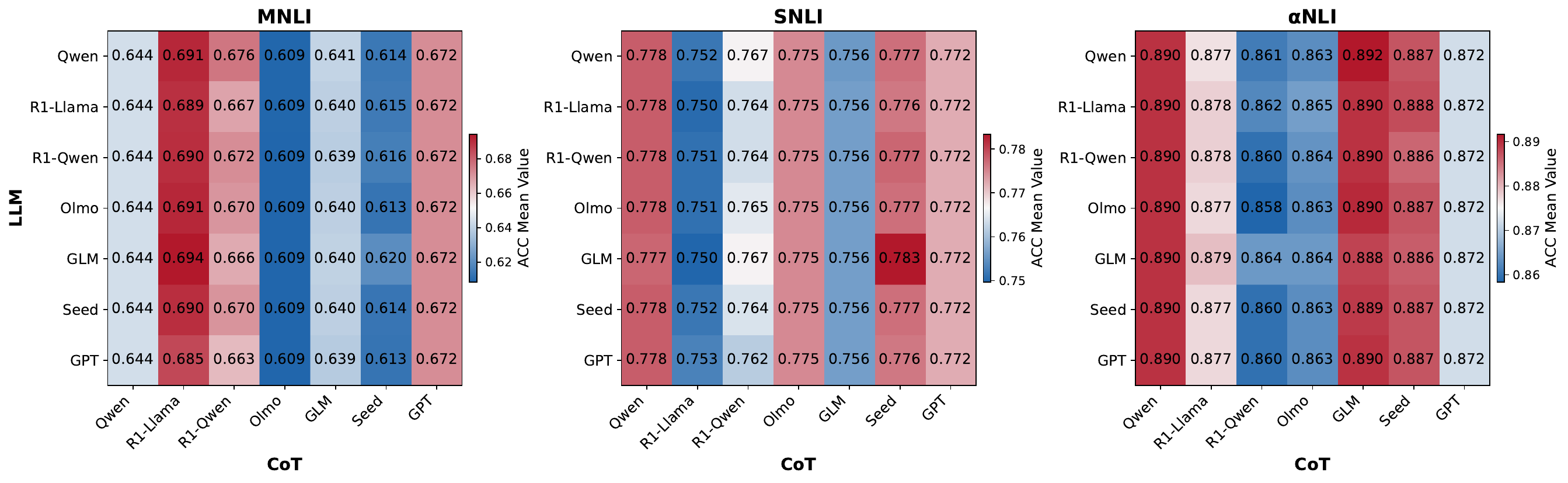}\label{fig:last-acc}
    }
    
    \subfigure[Last JSD ↓]{
        \includegraphics[width=0.95\linewidth]{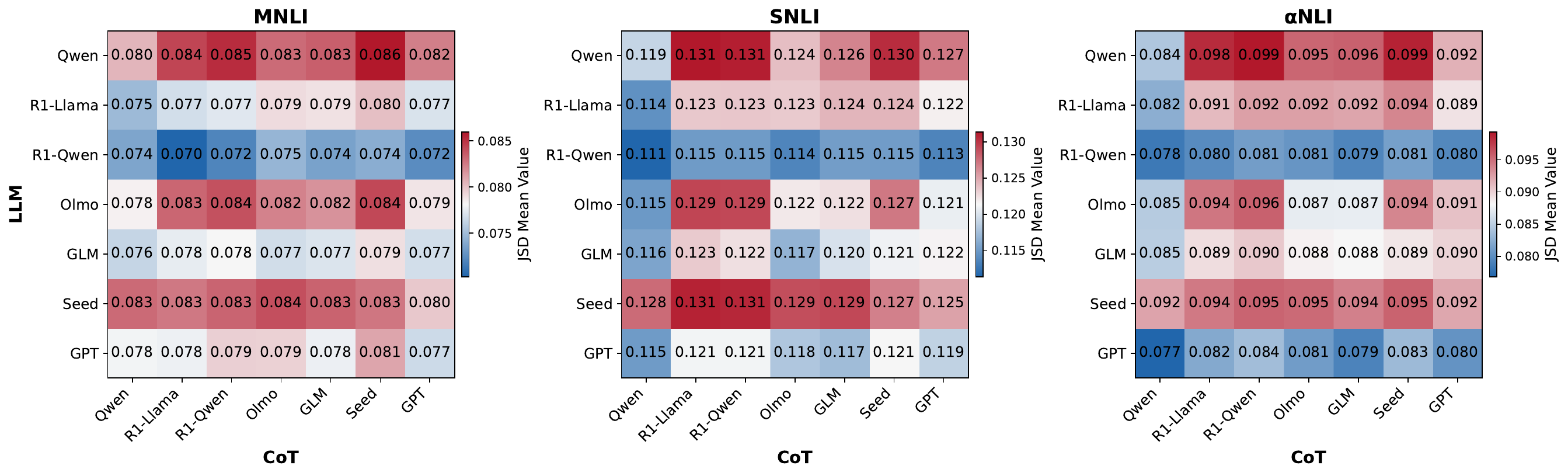}\label{fig:last-jsd}
    }

    \subfigure[Last Spearman’s $\rho$ ↑]{
        \includegraphics[width=0.95\linewidth]{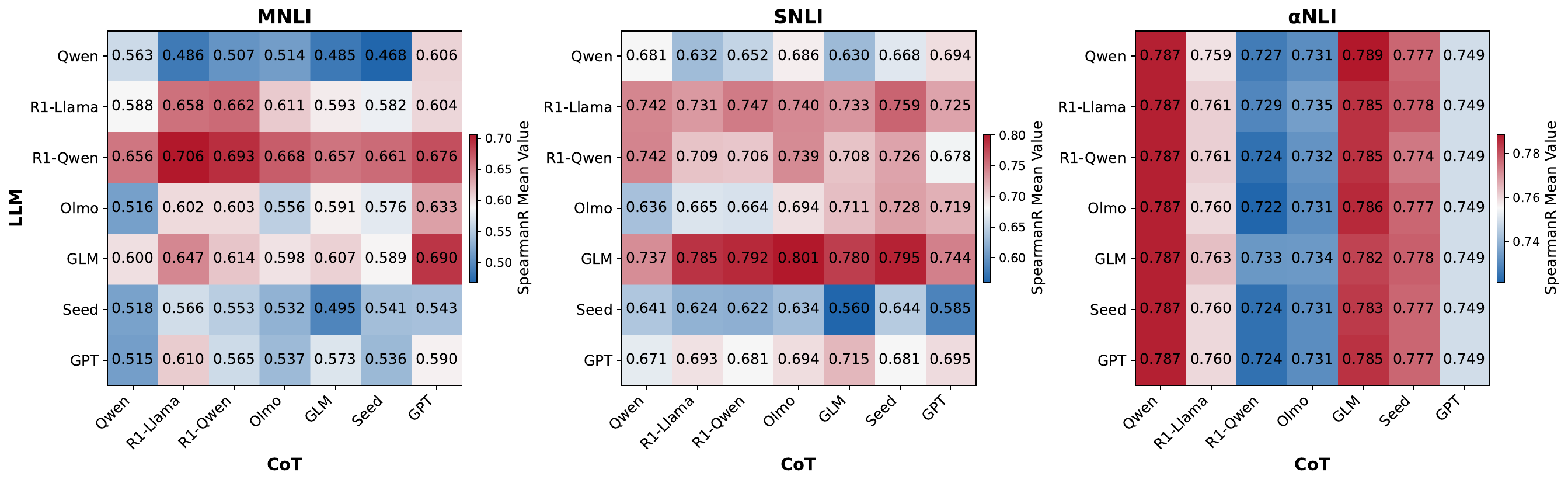}\label{fig:last-spearmanr}
    }
    \caption{Last results after reasoning for Cross-CoT.}
    \label{fig:last}
\end{figure}

\begin{table*}[t]
\centering
\resizebox{\linewidth}{!}{
\begin{tabular}{lccccccccccccccccccccccccccc}
\toprule
\textbf{Task}        & \multicolumn{9}{c}{MNLI}                                                                 & \multicolumn{9}{c}{SNLI}                                                                  & \multicolumn{9}{c}{$\alpha$NLI}                                                                  \\
\cmidrule(lr){1-1} \cmidrule(lr){2-10} \cmidrule(lr){11-19} \cmidrule(lr){20-28} 
\textbf{Metric}      & \multicolumn{3}{c}{ACC}     & \multicolumn{3}{c}{JSD}    & \multicolumn{3}{c}{Spearman’s $\rho$} & \multicolumn{3}{c}{ACC}     & \multicolumn{3}{c}{JSD}     & \multicolumn{3}{c}{Spearman’s $\rho$} & \multicolumn{3}{c}{ACC}     & \multicolumn{3}{c}{JSD}     & \multicolumn{3}{c}{Spearman’s $\rho$} \\
\cmidrule(lr){1-1} \cmidrule(lr){2-4} \cmidrule(lr){5-7} \cmidrule(lr){8-10} \cmidrule(lr){11-13} \cmidrule(lr){14-16} \cmidrule(lr){17-19} \cmidrule(lr){20-22} \cmidrule(lr){23-25} \cmidrule(lr){26-28}     
\textbf{Step/Factor} & LLM     & CoT    & Residual & LLM     & CoT   & Residual & LLM      & CoT     & Residual & LLM     & CoT    & Residual & LLM     & CoT    & Residual & LLM      & CoT     & Residual & LLM     & CoT    & Residual & LLM     & CoT    & Residual & LLM      & CoT     & Residual \\
\midrule
Step 0      & \hlc{lightred}{100.0\%} & 0.0\%  & 0.0\%    & \hlc{lightred}{100.0\%} & 0.0\% & 0.0\%    & \hlc{lightred}{100.0\%}  & 0.0\%   & 0.0\%    & \hlc{lightred}{100.0\%} & 0.0\%  & 0.0\%    & \hlc{lightred}{100.0\%} & 0.0\%  & 0.0\%    & \hlc{lightred}{100.0\%}  & 0.0\%   & 0.0\%    & \hlc{lightred}{100.0\%} & 0.0\%  & 0.0\%    & \hlc{lightred}{100.0\%} & 0.0\%  & 0.0\%    & \hlc{lightred}{100.0\%}  & 0.0\%   & 0.0\%    \\
Step 1      & \hlc{lightred}{97.3\%}  & 0.2\%  & 2.5\%    & \hlc{lightred}{96.9\%}  & 0.4\% & 2.7\%    & \hlc{lightred}{92.7\%}   & 0.4\%   & 6.9\%    & \hlc{lightred}{96.5\%}  & 1.4\%  & 2.2\%    & \hlc{lightred}{94.4\%}  & 0.9\%  & 4.7\%    & \hlc{lightred}{83.2\%}   & 2.4\%   & 14.4\%   & \hlc{lightred}{93.9\%}  & 1.4\%  & 4.7\%    & \hlc{lightred}{95.4\%}  & 1.3\%  & 3.3\%    & \hlc{lightred}{94.0\%}   & 1.3\%   & 4.8\%    \\
Step 2      & \hlc{lightred}{95.7\%}  & 0.5\%  & 3.7\%    & \hlc{lightred}{95.3\%}  & 0.7\% & 4.0\%    & \hlc{lightred}{92.3\%}   & 1.8\%   & 5.9\%    & \hlc{lightred}{85.1\%}  & 8.6\%  & 6.4\%    & \hlc{lightred}{88.2\%}  & 4.4\%  & 7.4\%    & \hlc{lightred}{76.1\%}   & 8.0\%   & 15.8\%   & \hlc{lightred}{75.7\%}  & 10.5\% & 13.8\%   & \hlc{lightred}{89.3\%}  & 3.9\%  & 6.8\%    & \hlc{lightred}{76.3\%}   & 10.4\%  & 13.2\%   \\
Step 3      & \hlc{lightred}{91.8\%}  & 1.6\%  & 6.6\%    & \hlc{lightred}{93.4\%}  & 1.6\% & 5.0\%    & \hlc{lightred}{91.6\%}   & 2.6\%   & 5.8\%    & \hlc{lightred}{75.1\%}  & 16.7\% & 8.1\%    & \hlc{lightred}{82.2\%}  & 8.3\%  & 9.5\%    & \hlc{lightred}{73.2\%}   & 12.2\%  & 14.6\%   & \hlc{lightred}{64.1\%}  & 22.2\% & 13.7\%   & \hlc{lightred}{81.4\%}  & 8.8\%  & 9.8\%    & \hlc{lightred}{64.3\%}   & 22.4\%  & 13.3\%   \\
Step 4      & \hlc{lightred}{89.1\%}  & 2.2\%  & 8.7\%    & \hlc{lightred}{91.5\%}  & 2.9\% & 5.6\%    & \hlc{lightred}{90.5\%}   & 3.6\%   & 5.9\%    & \hlc{lightred}{62.5\%}  & 29.1\% & 8.5\%    & \hlc{lightred}{75.4\%}  & 14.0\% & 10.6\%   & \hlc{lightred}{69.9\%}   & 16.4\%  & 13.7\%   & \hlc{lightred}{52.1\%}  & 31.2\% & 16.8\%   & \hlc{lightred}{80.8\%}  & 9.5\%  & 9.6\%    & \hlc{lightred}{51.9\%}   & 32.6\%  & 15.5\%   \\
Step 5      & \hlc{lightred}{85.2\%}  & 4.3\%  & 10.6\%   & \hlc{lightred}{89.3\%}  & 4.2\% & 6.4\%    & \hlc{lightred}{88.0\%}   & 4.9\%   & 7.1\%    & \hlc{lightred}{57.4\%}  & 32.4\% & 10.2\%   & \hlc{lightred}{72.8\%}  & 16.6\% & 10.6\%   & \hlc{lightred}{70.8\%}   & 13.0\%  & 16.2\%   & \hlc{lightred}{45.2\%}  & 41.7\% & 13.1\%   & \hlc{lightred}{81.6\%}  & 10.4\% & 8.0\%    & \hlc{lightred}{47.2\%}   & 41.3\%  & 11.5\%   \\
Step 6      & \hlc{lightred}{79.4\%}  & 8.5\%  & 12.0\%   & \hlc{lightred}{87.9\%}  & 5.7\% & 6.5\%    & \hlc{lightred}{83.6\%}   & 7.6\%   & 8.8\%    & \hlc{lightred}{58.9\%}  & 29.7\% & 11.3\%   & \hlc{lightred}{72.5\%}  & 17.4\% & 10.1\%   & \hlc{lightred}{71.9\%}   & 10.4\%  & 17.7\%   & 35.4\%  & \hlc{lightred}{53.8\%} & 10.8\%   & \hlc{lightred}{84.9\%}  & 9.4\%  & 5.8\%    & 36.4\%   & \hlc{lightred}{53.3\%}  & 10.3\%   \\
Step 7      & \hlc{lightred}{66.5\%}  & 18.4\% & 15.1\%   & \hlc{lightred}{87.5\%}  & 6.1\% & 6.4\%    & \hlc{lightred}{77.5\%}   & 10.1\%  & 12.4\%   & \hlc{lightred}{53.4\%}  & 30.6\% & 16.0\%   & \hlc{lightred}{73.1\%}  & 17.2\% & 9.7\%    & \hlc{lightred}{73.9\%}   & 6.6\%   & 19.5\%   & 22.5\%  & \hlc{lightred}{67.3\%} & 10.2\%   & \hlc{lightred}{87.0\%}  & 8.2\%  & 4.9\%    & 22.1\%   & \hlc{lightred}{67.5\%}  & 10.4\%   \\
Step 8      & \hlc{lightred}{44.1\% } & 38.3\% & 17.5\%   & \hlc{lightred}{86.4\%}  & 7.0\% & 6.7\%    & \hlc{lightred}{73.3\%}   & 14.5\%  & 12.2\%   & \hlc{lightred}{43.0\%}  & 33.2\% & 23.8\%   & \hlc{lightred}{72.1\%}  & 19.1\% & 8.8\%    & \hlc{lightred}{71.8\%}   & 7.9\%   & 20.3\%   & 9.6\%   & \hlc{lightred}{76.2\%} & 14.2\%   & \hlc{lightred}{85.2\%}  & 10.1\% & 4.7\%    & 10.1\%   & \hlc{lightred}{76.9\%}  & 13.0\%   \\
Step 9      & 22.4\%  & \hlc{lightred}{65.2\%} & 12.4\%   & \hlc{lightred}{84.3\%}  & 8.5\% & 7.2\%    & \hlc{lightred}{73.0\%}   & 16.8\%  & 10.3\%   & 34.8\%  & \hlc{lightred}{43.9\%} & 21.2\%   & \hlc{lightred}{71.3\%}  & 19.8\% & 8.9\%    & \hlc{lightred}{77.8\%}   & 6.3\%   & 15.9\%   & 5.3\%   & \hlc{lightred}{83.5\%} & 11.2\%   & \hlc{lightred}{81.4\%}  & 13.2\% & 5.4\%    & 5.4\%    & \hlc{lightred}{84.1\%}  & 10.5\%   \\
Step 10     & 0.1\%   & \hlc{lightred}{99.5\%} & 0.4\%    & \hlc{lightred}{83.3\%}  & 8.7\% & 8.1\%    & \hlc{lightred}{71.1\%}   & 13.1\%  & 15.7\%   & 0.2\%   & \hlc{lightred}{98.6\%} & 1.2\%    & \hlc{lightred}{67.7\%}  & 22.2\% & 10.1\%   & \hlc{lightred}{81.7\%}   & 3.3\%   & 15.0\%   & 0.1\%   & \hlc{lightred}{99.4\%} & 0.5\%    & \hlc{lightred}{76.9\%}  & 15.2\% & 7.9\%    & 0.1\%    & \hlc{lightred}{99.4\%}  & 0.5\%   \\
\bottomrule
\end{tabular}}
\caption{Step-wise ANOVA results. Each CoT is split into 10 segments by sentence, yielding 11 intermediate answers from no-thinking (step 0) to full-thinking (step 10). ANOVA is computed for each Cross-CoT heatmap. Red numbers indicate the factor dominating the metric at that step. All step-wise heatmaps are in Appendix~\ref{app:heatmaps}.}
\label{tab:anova}
\end{table*}

\begin{figure*}[t]

        \centering
        \includegraphics[width=0.9\linewidth]{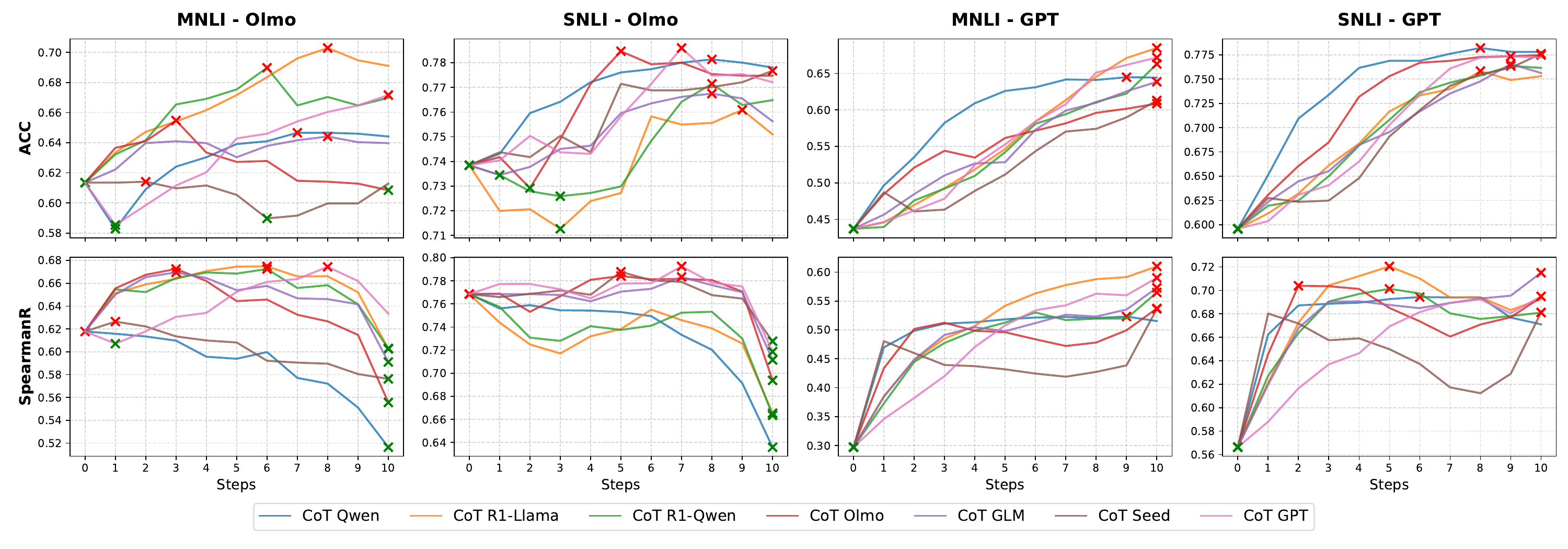}
        \caption{Curve cases for step-wise evaluation. Max and min points are marked. All results are in Appendix~\ref{app:curves}.}
        \label{fig:curves}
\end{figure*}

\section{Analyses}
\subsection{What Does CoT Determine?}
\label{sec:Last_CoT}

If CoT were the dominant driver, models conditioned on the same CoT should converge to similar outcomes. To test this, we analyze final-step accuracy and JSD.
The results (Figure~\ref{fig:last-acc},~\ref{fig:last-jsd}) reveal a clear dissociation. Accuracy shows a \textit{column-dominant pattern}: for a fixed CoT source, accuracy is nearly identical across inference models, indicating that CoT largely dictates the argmax decision. In contrast, JSD exhibits a \textit{row-dominant pattern}: final divergence is primarily determined by the inference model, with little sensitivity to the CoT. 
This suggests that CoT provides a directional signal, while the final distributional shape remains constrained by model-specific priors.

We further analyze Spearman’s $\rho$ as a relaxed non-argmax metric (only rankings). Its heatmap (Figure~\ref{fig:last-spearmanr}) mirrors the row-dominant structure of JSD rather than accuracy.\!\footnote{Note that, in binary-choice $\alpha$NLI, $\rho$ is effectively equivalent to accuracy, as only a single non-argmax position exists.} We adopt an additive Analysis of Variance (ANOVA) model to estimate marginal contributions of CoT and the model's priors. 
Interaction effects, while not explicitly modeled, are captured in the residual term.\!\footnote{Details of ANOVA are in Appendix~\ref{app:anova}.} 
ANOVA on MNLI confirms this split: CoT explains 99\% of the variance in accuracy, but only a small fraction in JSD and Spearman’s $\rho$ (8.7\% and 13.1\%), where model identity dominates (83.3\% and 71.1\%).
This asymmetry exposes a fundamental limitation of current CoT paradigms. CoT is highly effective at explicit decision-making, capable of overriding a model’s prior to determine the argmax. However, in the non-argmax space—namely, the ranking and probability allocation over alternative options—its influence sharply diminishes. \textbf{Models appear to follow CoT for the final choice, but revert to their latent parametric preferences when distributing uncertainty.}

\subsection{When Does CoT Take Control?}
\label{sec:when}

Although CoT explains nearly all variance in accuracy, its impact on distributional metrics remains limited. 
To understand how this asymmetry develops, we apply early stopping to the CoT, truncating it at fixed increments and evaluating intermediate performance.\!\footnote{Implementation details are elaborated in the Appendix~\ref{app:early-stop}. All code, logits, and CoT outputs are publicly available at \href{https://github.com/mainlp/CoT-HLV}{https://github.com/mainlp/CoT-HLV}. }
Step-wise ANOVA (Table~\ref{tab:anova}) reveals a sharp divergence. CoT influence on accuracy remains modest during reasoning, then spikes abruptly at the final step, forming a clear inflection. In contrast, its influence on JSD and Spearman’s $\rho$ stays uniformly low, with no point at which CoT overrides the model’s ranking behavior.
To ensure our findings are not artifacts of the chosen probability mapping, we additionally compute JSD using a standard softmax transformation. As detailed in Appendix~\ref{app:robust}, this robustness check yields consistent results, confirming the decoupled mechanism regardless of the normalization scheme.

Notably, the residual variance in our ANOVA decomposition remains consistently small (often under 10\%). This indicates that the main effects overwhelmingly dominate the predictive variance. Because the additive ANOVA model absorbs any unmodeled interactions into the residual term, the minimal magnitude of these residuals assures that the primary conclusions regarding the decoupled mechanism remain robust.

This pattern is also illustrated by representative models (Figure~\ref{fig:curves}). Accuracy often shifts or converges only at the conclusion, while Spearman’s $\rho$ fluctuates without a consistent trend. Thus, CoT determines the LLM's final choice but not the structure of uncertainty.
Our anecdotal evidence in Appendix~\ref{sec:appendix_qualitative} supports that this can be attributed to the CoT format:
standard CoT often ends with an explicit conclusion, providing a strong argmax signal, while distributional cues remain implicit.
Consequently, \textbf{models leverage CoT for decision-making but revert to intrinsic priors for probability allocation, exposing a structural inability of raw CoT to shape answer distributions.}

\section{Discussion and Future Work}

Our results indicate that scaling inference-time compute via longer CoT is insufficient for resolving complex semantic ambiguity. Rather than preserving or refining uncertainty, standard CoT primarily acts as a mechanism for collapsing a distribution over hypotheses into a single high-confidence argmax prediction. This suggests a broader limitation of current reasoning paradigms: they are primarily optimized for decisiveness, rather than for faithfully representing uncertainty.

Beyond motivating the need for improved HLV evaluation, our findings point to several concrete directions for advancing both modeling and analysis of reasoning systems.

First, to better capture HLV, future work should move beyond implicit reasoning traces and develop explicitly distribution-aware reasoning frameworks. Instead of solely optimizing for the final answer, models should be trained to maintain, calibrate, and communicate relative uncertainties over competing hypotheses throughout intermediate reasoning steps. This may involve new training objectives, decoding strategies, or supervision signals that explicitly reward distributional fidelity.

\paragraph{Scope of Evaluation.}
Our study focuses on NLI-style multiple-choice question answering (MCQA), a deliberately controlled setting that enables reliable extraction of probability distributions via first-token probabilities. This design avoids the additional noise introduced by open-ended generation, such as answer normalization and semantic equivalence issues. While MCQA is a standard benchmark for evaluating reasoning in LLMs, extending our distributional analysis to settings with larger label spaces and open-ended outputs is a natural next step. Such extensions would require more robust methods for mapping free-form generations to structured belief distributions.

\paragraph{Calibration vs.\ Distributional Similarity.}
Our evaluation framework emphasizes relative distributional alignment, measuring how closely model-predicted belief distributions match human annotations using metrics such as JSD and Spearman's $\rho$. However, this setup does not assess absolute probability calibration with respect to empirical correctness frequencies. A promising direction for future work is to jointly study distributional similarity and calibration, for example by incorporating metrics such as Expected Calibration Error (ECE), to better understand whether reasoning models are not only structurally aligned with human beliefs but also probabilistically well-grounded.

\paragraph{CoT Format Ablations.}
Finally, while the observed decoupling effect remains consistent across different source and reasoning models in our Cross-CoT framework, we do not isolate the causal contribution of specific reasoning formats. In particular, it remains unclear whether elements such as explicit final-answer cues or structural regularities in CoT directly govern decoding behavior. Future work could investigate this question through controlled format-level ablations, synthetic CoT interventions, or human-authored reasoning traces, enabling a more fine-grained understanding of how reasoning structure influences model predictions.

\section{Conclusion}

From an HLV perspective, we identify a fundamental ``split influence'' in CoT reasoning: while reasoning content predominantly determines the LLM's final argmax choice latently, the probability landscape of alternative options remains anchored to the model's intrinsic priors. This exposes a structural limitation where standard CoT effectively collapses ambiguity for decision-making but fails to calibrate fine-grained uncertainty for alternative, plausible answers. Consequently, advancing HLV modeling requires moving beyond implicit reasoning traces toward distribution-aware paradigms.

\section*{Limitations}

Our work has two main limitations.
First, our evaluation relies solely on final human label distributions, as ChaosNLI lacks annotated intermediate reasoning steps. Consequently, our step-wise analysis compares intermediate model outputs against the final human consensus rather than step-specific ground truth. Addressing this limitation would require future improvements in human annotation, where reasoning steps and intermediate answers are collected for direct comparison. An alternative approach could be the use of relative references: for example, treating the intermediate answers generated by a CoT-provider LLM as the gold standard to evaluate the faithfulness of other LLMs. Another possibility is to employ an entailment model to determine whether each reasoning step in a CoT entails the previous step, thereby inferring intermediate answers recursively. However, both of these approaches are highly dependent on the accuracy of the model itself; inaccuracies could introduce evaluation biases.

Second, our study does not include a direct human evaluation of the textual content of the CoTs. Instead, we focus on assessing CoTs in terms of their impact on LLM behavior, especially answer distributions. While this approach emphasizes the effect of reasoning on model outputs, it overlooks the quality of the specific text content. Conducting human evaluation of long CoTs is particularly resource-intensive, given their length and the effort required to annotate individual sentences and their interrelations. Nevertheless, considering the growing importance of CoT reasoning in NLP, carefully designed human evaluation to verify whether extreme values in reasoning metrics correspond to reasonable positions in the text represents a promising direction for future work. Such evaluation could help us better understand the effects of CoT on LLM reasoning.

\section*{Ethical Considerations}
This work primarily involves the analysis of NLI datasets and open-sourced LLMs. All data used are publicly available and do not contain personally identifiable information. No sensitive or potentially harmful content is generated or utilized in this study. Therefore, we do not anticipate any ethical concerns arising from our work.

\paragraph{Use of AI Assistants}
The authors acknowledge the use of ChatGPT solely for correcting grammatical errors, enhancing the coherence of the final manuscript.

\section*{Acknowledgements}
We thank the members of the MaiNLP lab for their insightful feedback on earlier drafts of this paper. 
We specifically appreciate the suggestions of Siyao Peng, Jana Grimm, Florian Eichin and Benedetta Muscato. 
We are also grateful to the anonymous reviewers for their constructive feedback.
BC acknowledges his membership in the European Laboratory for Learning and Intelligent Systems (ELLIS) PhD program.
BP and RL are supported by ERC Consolidator Grant DIALECT 101043235.
AK is supported by the UK Research and Innovation (UKRI) Frontier Research Grant EP/Y031350/1 EQUATE (the UK government's funding guarantee for ERC Advanced Grants).

\appendix

\section{Datasets}
\label{app:dataset}

To evaluate the model's ability to capture collective human uncertainty and label disagreement, we utilize the \textbf{ChaosNLI} dataset \citep{nie-etal-2020-learn}. Unlike standard NLI benchmarks that typically rely on a single ``gold'' label derived from a majority vote (often among 3--5 annotators), ChaosNLI provides a dense distribution of human annotations.

\begin{itemize}
    \item \textbf{Data Source:} The dataset consists of purely English examples, selected from of SNLI (1514 items, \citealt{bowman-etal-2015-large}), MNLI (1599 items, \citealt{williams-etal-2018-broad}) and $\alpha$NLI (1532 items, \citealt{DBLP:conf/iclr/BhagavatulaBMSH20}).
    \item \textbf{Selection Criteria:} The examples were specifically chosen to target ambiguous instances. The authors filtered for examples where the original annotators disagreed (e.g., a 3 vs. 2 vote split) or where the model predictions significantly deviated from the majority label.
    \item \textbf{Annotation Process:} Each example in ChaosNLI is annotated by a crowd of $N=100$ independent workers. This high volume of annotators allows for the estimation of a true label distribution $y_{\text{human}}$ over the three classes (Entailment, Neutral, Contradiction), rather than a deterministic class label. As for $\alpha$NLI, annotators are asked to select the better hypothesis from a sentence pair for a given observation pair.
    \item \textbf{Objective:} The dataset serves as a testbed for measuring how well a model's predicted probability distribution $p_{\text{model}}$ aligns with the distribution of human judgment $y_{\text{human}}$, often measured via Jensen-Shannon Divergence (JSD).
\end{itemize}

\section{Evaluation Details}
\label{app:evaluation}

This section elaborates on the details of the experimental setup. 
We first describe the experiment details, and then introduce how the NLI task is transformed into a multiple-choice question answering (MCQA) format. We finally introduce the procedure for extracting and converting first-token probabilities, followed by a formal definition of the evaluation metrics used in this paper.

\subsection{Experiment Details}

All LLMs are evaluated using the initial or recommended parameter settings provided by their respective developers, ensuring that each model generates Chain-of-Thought (CoT) outputs consistent with its intended behavior and style.  
Since our analysis focuses on the model logits rather than the sampled textual outputs, variations in sampling-related parameters (e.g., temperature, top-$k$, or top-$p$) do not affect the logits-based evaluations.  
This design ensures that our comparisons reflect the models' intrinsic preference distributions rather than stochastic differences introduced by the decoding process.

All experiments were conducted on two NVIDIA A100-SXM4-80GB GPUs.  
On average, completing a single experiment—which involves generating step-wise intermediate logits for one LLM on a single dataset using a given Chain-of-Thought (CoT)—takes approximately 20 hours.  
This reflects the computational demands of step-wise evaluation across multiple inference steps and highlights the resource-intensive nature of detailed logit-level analyses for large language models.

\subsection{MCQA Format}
The conversion to the MCQA format is illustrated in Table~\ref{tab:app1-task-prompt}. Since MNLI and SNLI belong to the same category of NLI datasets, they are transformed using an identical three-way multiple-choice formulation. In contrast, $\alpha$NLI is converted using a separate binary-choice MCQA format, reflecting its distinct label structure.

\newcolumntype{P}[1]{>{\raggedright\arraybackslash}p{#1}}
\begin{table*}[t]
\centering
\resizebox{\textwidth}{!}{
\begin{tabular}{P{0.15\textwidth}|P{0.85\textwidth}}
\toprule 
\textbf{Datasets} & \textbf{MCQA Transformation} \\
\midrule
MNLI \& SNLI & Please determine whether the following statement is true (entailment), undetermined (neutral), or false (contradiction) given the context below and select ONE of the listed options and start your answer with a single letter. \newline Context: \{premise\} \newline Statement: \{hypothesis\} \newline A. Entailment \newline B. Neutral \newline C. Contradiction \newline Answer: \\ \midrule
$\alpha$NLI &  Please determine which of the two hypotheses (A or B) is more likely to explain the transition from the beginning observation to the ending observation and select ONE of the listed options and start your answer with a single letter. \newline Beginning: \{begining-observation\} \newline Ending: \{ending-observation\} \newline A. \{hypothesis1\} \newline B. \{hypothesis2\} \newline Answer:     \\
\bottomrule
\end{tabular}}
\caption{
The MCQA transformation for NLI tasks.
}
\label{tab:app1-task-prompt}
\end{table*}

\subsection{First-Token-Probability and Metrics}

\subsubsection{First-token Probability}

Take MNLI as an example.
Conditioned on the prompts described above, we further map LLM outputs from discrete options in \texttt{[A, B, C]} to probability distributions, which we treat as model judgment distributions (MJDs).
Specifically, we define a one-to-one mapping
$f \colon O \rightarrow L$ from the option set $O$ to the label space $L$, where $O = \{A, B, C\}$ and $L = \{\textsc{Entailment}, \textsc{Neutral}, \textsc{Contradiction}\}$.
Both $O$ and $L$ are subject to permutation to mitigate positional and label-order biases.

Let the textual output of an LLM be represented as a sequence of tokens
$\bm{w} = [w_1, w_2, \dots, w_k]$, where $w_i \in V$, $k$ denotes the output length, and $V$ is the model vocabulary.
Instead of using the decoded output, we extract the pre-decoding logits corresponding to the first generated token $w_1$:
\[
\bm{s}_{w_1} = [s_1, s_2, \dots, s_n], \quad n = |V|,
\]
where $s_j$ denotes the logit associated with the $j$-th vocabulary token.

We restrict our attention to the subset of logits corresponding to the option tokens in $O$,
\[
\bm{s}^{O}_{w_1} = [s_A, s_B, s_C],
\]
which encode the model’s relative preference over the candidate options.
Since the normalization transformation preserves the entropy of the original logits, whereas the softmax transformation (especially when applied with a temperature parameter) can alter entropy, we adopt the normalization transformation for our evaluations, because entropy plays a critical role in the computation of JSD, and using a transformation that artificially modifies it could bias the assessment.  
Therefore, to more accurately measure the LLMs’ intrinsic probabilistic preferences and their native reasoning behavior, we rely on the norm transformation rather than softmax in our analysis.

To convert these scores into a probability distribution $\bm{p}^{O}$, we then apply a normalization step.

\begin{equation}
p^{O}_{\text{norm}}(j) = \frac{s_j}{\sum_{j=1}^{|O|} s_j},
\end{equation}

This procedure yields a well-formed probability distribution over labels, enabling fine-grained comparison with human-annotated label distributions.

\subsubsection{Rank Correlation Metric}
To quantify the agreement between ranked preferences from different sources (e.g., human annotations versus model predictions), we employ rank correlation metrics.
Let $\{(x_i, y_i)\}_{i=1}^{n}$ denote paired ranks from two sources.

\paragraph{Spearman’s $\rho$} \cite{spearman1961proof}
Spearman’s rank correlation coefficient measures the Pearson correlation between ranked variables and is defined as:
\begin{equation}
\rho = 1 - \frac{6 \sum_{i=1}^{n} d_i^2}{n(n^2 - 1)} ,
\end{equation}
where $d_i = x_i - y_i$ denotes the rank difference for the $i$-th item.
Spearman’s $\rho$ captures monotonic relationships and is robust to nonlinear transformations of the underlying scores.

\subsubsection{Distribution-Based Metric}
For settings where both human annotations and model outputs are represented as probability distributions, we adopt distributional similarity metrics.

\paragraph{Jensen--Shannon Distance (JSD)} \cite{DBLP:journals/tit/EndresS03}
Given two discrete probability distributions $P$ and $Q$, the Jensen--Shannon Distance is defined as:
\begin{equation}
D_{\text{JSD}}(P \| Q) =
\sqrt{\frac{1}{2} \left(
D_{\text{KL}}(P \| M) + D_{\text{KL}}(Q \| M)
\right)} ,
\end{equation}
where $M = \frac{1}{2}(P + Q)$ and $D_{\text{KL}}(\cdot \| \cdot)$ denotes the Kullback--Leibler divergence.
JSD is symmetric, bounded, and well-defined even when $P$ and $Q$ contain zero-probability entries, making it suitable for comparing soft label distributions.

\section{Box-plot for Cross-CoT Experiments}
\label{app:Cross-CoT}
\begin{figure*}[t]

        \centering
        \includegraphics[width=\linewidth]{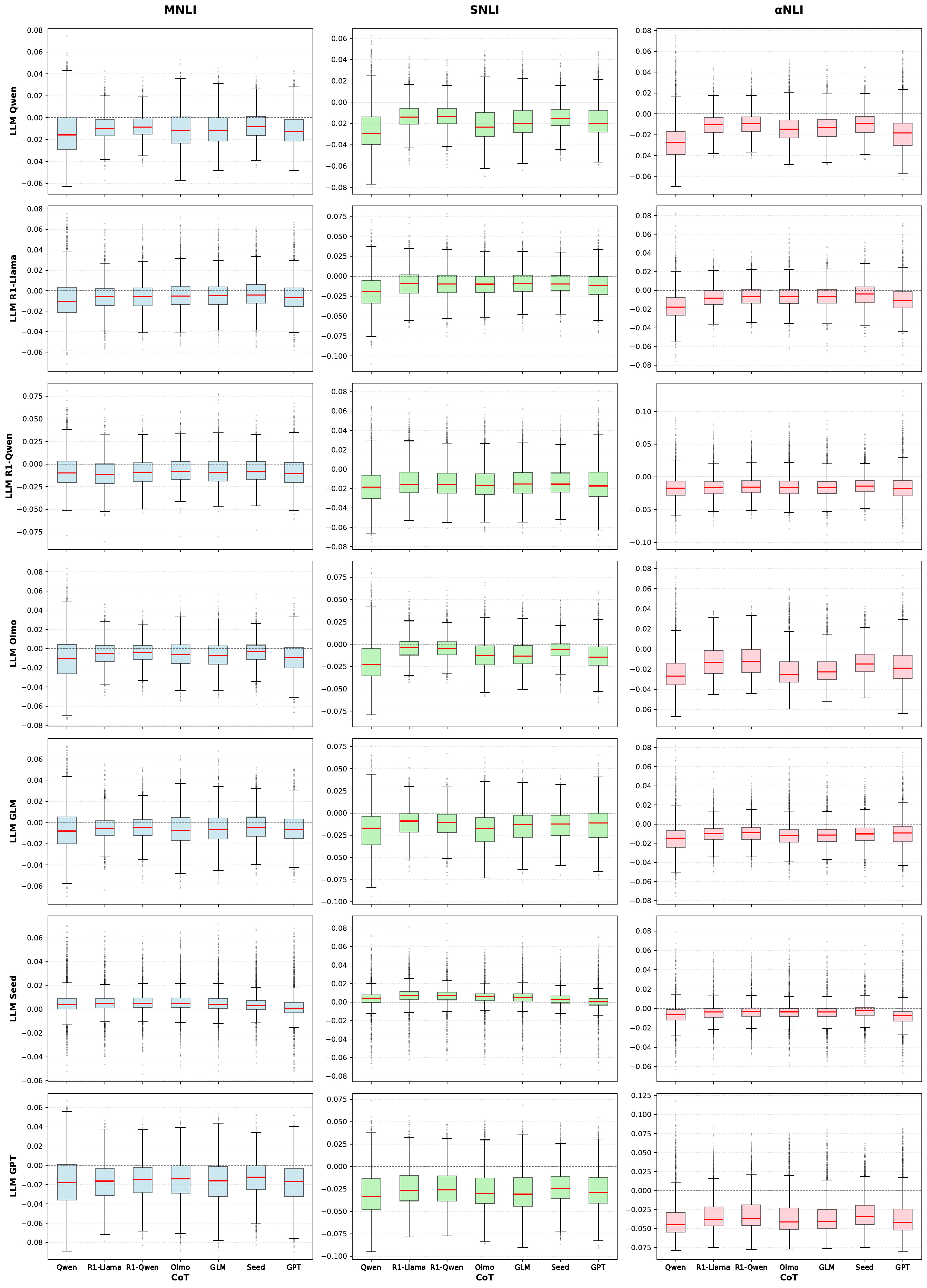}
        \caption{Delta JSD box plot.}
        \label{fig:delta-jsd-box}
\end{figure*}

This section presents the distribution of the Delta JSD metric over all instances in the dataset, aiming to show that the observed reduction in JSD reflects a global trend rather than being driven by a small number of extreme cases that artificially lower the mean.

As illustrated in Figure~\ref{fig:delta-jsd-box}, we visualize per-instance Delta JSD values using box plots, where the central line denotes the median, the boxes correspond to the interquartile range (IQR), and the whiskers extend to $1.5\times$IQR.
Across the majority of experimental settings, the distributions are centered below zero, indicating a consistent overall decrease in JSD.
These results suggest that the improvement captured by the average JSD is broadly shared across data points, rather than being dominated by a few outliers.

\section{ANOVA Details}
\label{app:anova}

To quantify the relative influence of different factors on the observed scores, we employ a two-way Analysis of Variance (ANOVA) with an additive (no-interaction) design. This statistical framework allows us to decompose the total variance of the dependent variable into contributions from multiple categorical factors.

\subsection{Problem Setup}

Let $y_{ij}$ denote the observed score associated with the $i$-th model configuration and the $j$-th CoT setting. In our implementation, we consider:
\begin{itemize}
    \item A \textbf{model factor} with $I=7$ levels (indexed by $i$),
    \item A \textbf{CoT factor} with $J=7$ levels (indexed by $j$),
    \item One observation for each $(i,j)$ combination.
\end{itemize}

The data are organized into a long-form table with three columns:
\texttt{score}, \texttt{model}, and \texttt{param}, where both \texttt{model} and \texttt{param} are treated as categorical variables.

\subsection{Additive Two-Way ANOVA Model}

We adopt an additive two-way ANOVA model without interaction terms, formulated as:
\begin{equation}
y_{ij} = \mu + \alpha_i + \beta_j + \varepsilon_{ij},
\end{equation}
where:
\begin{itemize}
    \item $\mu$ is the overall mean score,
    \item $\alpha_i$ represents the main effect of the $i$-th model,
    \item $\beta_j$ represents the main effect of the $j$-th CoT,
    \item $\varepsilon_{ij}$ is the residual error term.
\end{itemize}

This formulation assumes that the effects of the two factors are independent and additive, and that no interaction between model and CoT is modeled. This choice aligns with our goal of isolating the marginal contribution of each factor.

\subsection{Estimation via Ordinary Least Squares}

The model is estimated using Ordinary Least Squares (OLS), implemented as:
\begin{equation}
\texttt{score} \sim C(\texttt{model}) + C(\texttt{param}),
\end{equation}
where $C(\cdot)$ denotes categorical encoding. The fitted model is then passed to a Type-II ANOVA procedure, which computes sums of squares for each main effect after accounting for the other factor.

\subsection{Variance Decomposition}

ANOVA decomposes the total sum of squares (SS) as:
\begin{equation}
SS_{\text{total}} = SS_{\text{model}} + SS_{\text{param}} + SS_{\text{residual}},
\end{equation}
where:
\begin{itemize}
    \item $SS_{\text{model}}$ captures variance explained by the model factor,
    \item $SS_{\text{param}}$ captures variance explained by the CoT factor,
    \item $SS_{\text{residual}}$ captures unexplained variance.
\end{itemize}

Each sum of squares is associated with an $F$-statistic and corresponding $p$-value, allowing statistical significance testing of factor effects.

\subsection{Contribution Percentage}

To improve interpretability, we further compute the \textbf{variance contribution percentage} of each factor:
\begin{equation}
\text{Contribution}_k = 
\frac{SS_k}{SS_{\text{total}}} \times 100\%,
\end{equation}
where $k \in \{\text{model}, \text{param}, \text{residual}\}$.

This metric reflects the proportion of total variance attributable to each source. In our implementation, these percentages are rounded to one decimal place and reported as the final output of the analysis.

\subsection{Interpretation}

The resulting contribution percentages provide a clear quantitative comparison of how much variability in the scores is explained by:
\begin{itemize}
    \item differences between models,
    \item differences between CoT settings,
    \item unexplained residual noise.
\end{itemize}

This two-way additive ANOVA thus serves as an effective tool for disentangling and comparing the marginal effects of multiple experimental factors in our evaluation framework.

Note that our analysis employs an additive two-way ANOVA without explicit interaction terms. Since LLM inference in our framework is fully deterministic—yielding identical pre-decoding logits for a fixed prompt and CoT sequence—multiple runs per experimental cell are unnecessary. Consequently, the additive model is specifically designed to isolate and quantify the marginal effects of the CoT content versus the intrinsic model prior. Any potential interaction effects, along with unexplained variance, are cleanly absorbed into the residual error term ($\epsilon_{ij}$).

\section{Implementation for Early Stopping}
\label{app:early-stop}

\subsection{Accumulative 10\% Segmenting of Chain-of-Thoughts}

To facilitate analysis of reasoning progression in CoT outputs, we segment each text into ten accumulative portions, corresponding approximately to every 10\% of the text length~\cite{DBLP:journals/corr/abs-2509-14004,zhao-etal-2026-comprehensive,DBLP:journals/corr/abs-2601-02996}. Formally, given a text $T$ of length $L$, we aim to identify cut points $p_1, p_2, \dots, p_9$ such that $p_i$ roughly corresponds to $i \cdot L / 10$, with the final point $p_{10} = L$. 

Our procedure combines sentence-aware and heuristic splitting strategies. First, we parse $T$ into sentences using a syntactic parser and extract all sentence-ending positions. If at least nine sentences are available, we select each cut point $p_i$ as the nearest sentence end to $i \cdot L / 10$, ensuring monotonicity of cut points. If fewer than ten sentences exist, we iteratively split the longest existing segment, prioritizing natural boundaries such as punctuation (e.g., semicolons, commas) and spaces, and resorting to midpoint splits when no suitable boundary is found. 

Finally, we enforce strict monotonicity of cut points and construct the accumulative segments $S_1, S_2, \dots, S_{10}$ where $S_i = T[:p_i]$. This ensures that the last segment always reproduces the full original text. This method preserves original spacing and punctuation, providing natural and interpretable checkpoints for CoT analysis at decile intervals.

\subsection{Early-Stopping and Answer Token Extraction}

To reliably extract intermediate reasoning outputs, we adopt an \emph{early-stopping} strategy. Specifically, at each accumulative CoT segment cut point, we append a special token sequence signaling the model to terminate reasoning and produce an answer, e.g., ``\textbackslash\texttt{n</think>}\textbackslash\texttt{n}\textbackslash\texttt{nBased on the reasoning so far, the Answer is:}''. This encourages the model to emit the \texttt{Answer} token at that intermediate stage, preventing incomplete or excessively long continuations. 

After obtaining the logits for the first token following this prompt, we convert them into a probability distribution using the \emph{first-token probability} method. This approach allows us to quantify the model's intermediate answer distribution at each 10\% reasoning checkpoint, providing a fine-grained view of decision-making evolution along the CoT.

\section{Robustness Check for the Softmax Transformation}
\label{app:robust}

Using logits (or first-token probabilities) as confidence proxies is standard in prior work on LLM uncertainty and MCQA evaluation~\cite{chen-etal-2024-seeing,DBLP:journals/corr/abs-2601-06407}. Because first-token scores are deterministic given the input and unaffected by sampling hyperparameters, they provide a stable signal of model preference, which our work follows.

Our evaluation focuses on the relative allocation of probability mass across options (Human Label Variation). \emph{We note that softmax introduces an exponential rescaling that may change the entropy of the original score distribution, while entropy is an important factor in uncertainty evaluation}. To better preserve the relative allocation structure across options and to maintain a more pronounced difference, we therefore adopt linear normalization over the candidate option tokens (A/B/C).

Importantly, in our MCQA setting we empirically observe that the logits at the A/B/C positions are consistently positive and well separated across all evaluated reasoning models and prompts—typically above 10. This avoids potential sign issues in linear normalization. We believe this property is partly due to our strongly constrained output format. Under these conditions, linear normalization yields valid non-negative vectors that sum to 1.

We also consider that softmax is a reasonable alternative and include it as an additional robustness check here. Robust check results (MNLI Step-ANOVA with softmax (t=10) JSD, Table~\ref{tab:robust}) below lead to the same conclusions as the linear-normalization JSD results reported in Table~\ref{tab:anova}.

\begin{table}[ht]
\centering
\begin{tabular}{cccc}
\toprule
Step & LLM (\%) & CoT (\%) & Residual (\%) \\
\midrule
0  & 100.0 & 0.0  & 0.0 \\
1  & 97.9  & 0.4  & 1.7 \\
2  & 96.7  & 0.6  & 2.6 \\
3  & 95.6  & 1.3  & 3.2 \\
4  & 94.3  & 1.9  & 3.8 \\
5  & 92.8  & 2.3  & 4.9 \\
6  & 91.4  & 3.1  & 5.4 \\
7  & 90.4  & 3.2  & 6.4 \\
8  & 88.8  & 3.8  & 7.4 \\
9  & 86.3  & 5.6  & 8.1 \\
10 & 77.1  & 13.3 & 9.6 \\
\bottomrule
\end{tabular}
\caption{Robust check for step-wise ANOVA on MNLI Softmax (t=10) JSD.}
\label{tab:robust}
\end{table}

\section{All Heatmaps for Step-wise Evaluation}
\label{app:heatmaps}

\begin{figure}[t]

        \centering
        \includegraphics[width=0.8\linewidth]{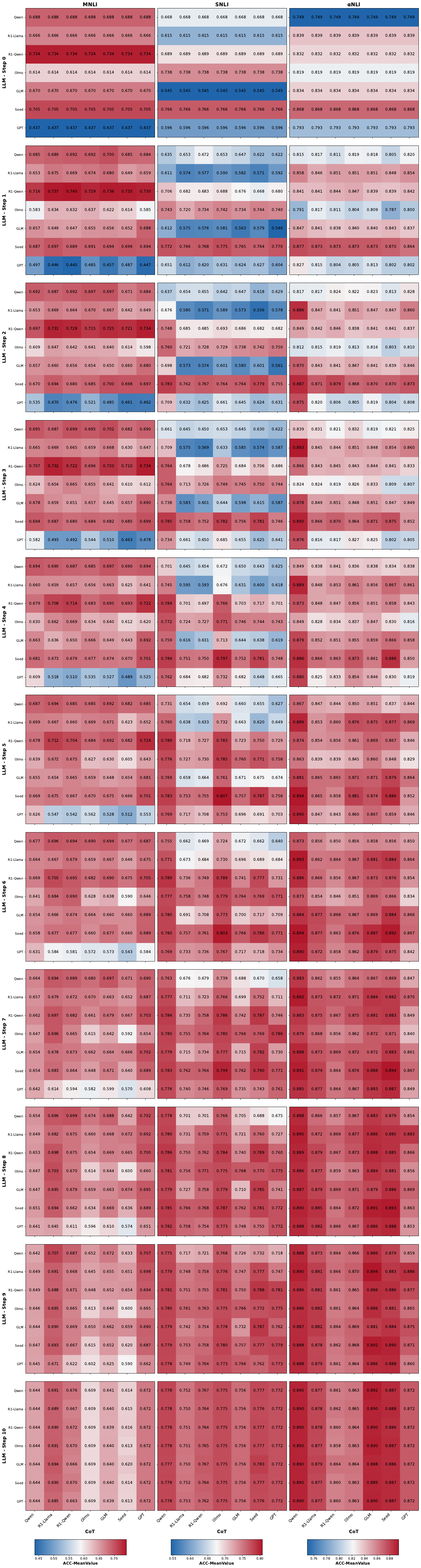}
        \caption{Steps ACC.}
        \label{fig:steps-acc}
\end{figure}

\begin{figure}[t]

        \centering
        \includegraphics[width=0.8\linewidth]{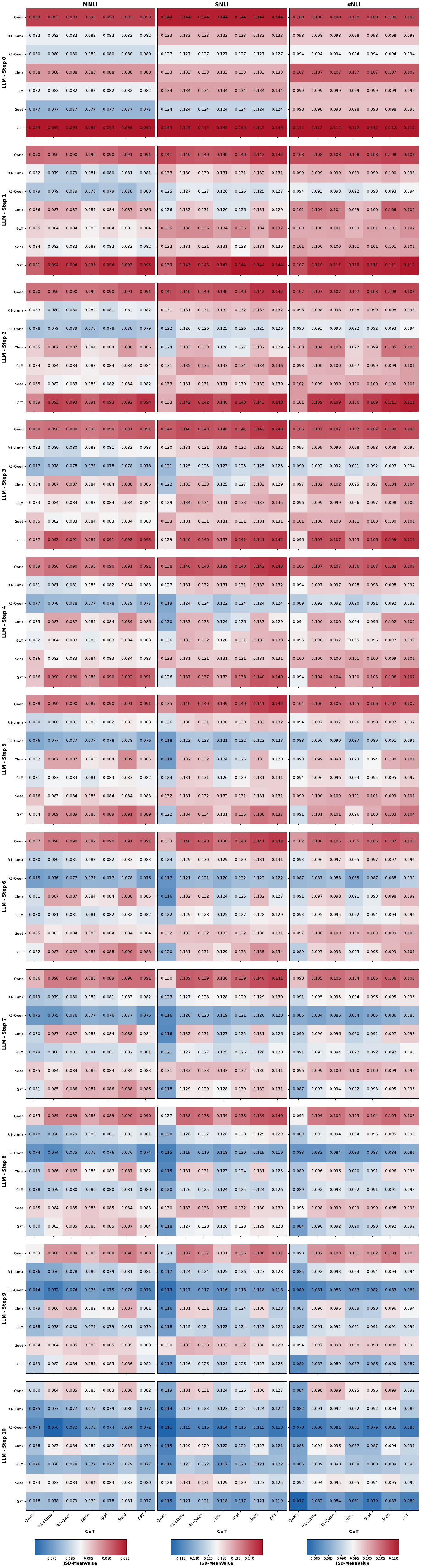}
        \caption{Steps JSD.}
        \label{fig:steps-jsd}
\end{figure}

\begin{figure}[t]

        \centering
        \includegraphics[width=0.8\linewidth]{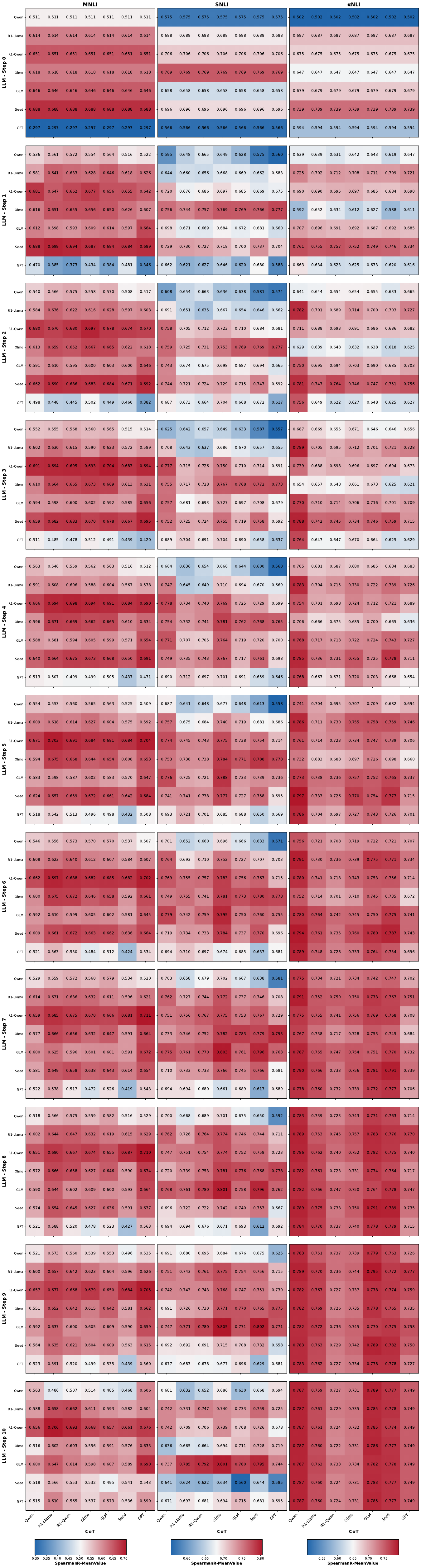}
        \caption{Steps Spearman’s $\rho$.}
        \label{fig:steps-spearmanr}
\end{figure}

This section presents the full set of heatmaps obtained from our step-wise evaluation, which are subsequently used to compute the ANOVA effect sizes reported in Table~\ref{tab:anova}. 
Specifically, we provide heatmaps for \textbf{accuracy} (Figure~\ref{fig:steps-acc}), \textbf{JSD} (Figure~\ref{fig:steps-jsd}), \textbf{Spearman’s $\rho$} (Figure~\ref{fig:steps-spearmanr}).

\section{All Curves for Step-wise Evaluation}
\label{app:curves}

\begin{figure*}[t]

        \centering
        \includegraphics[width=\linewidth]{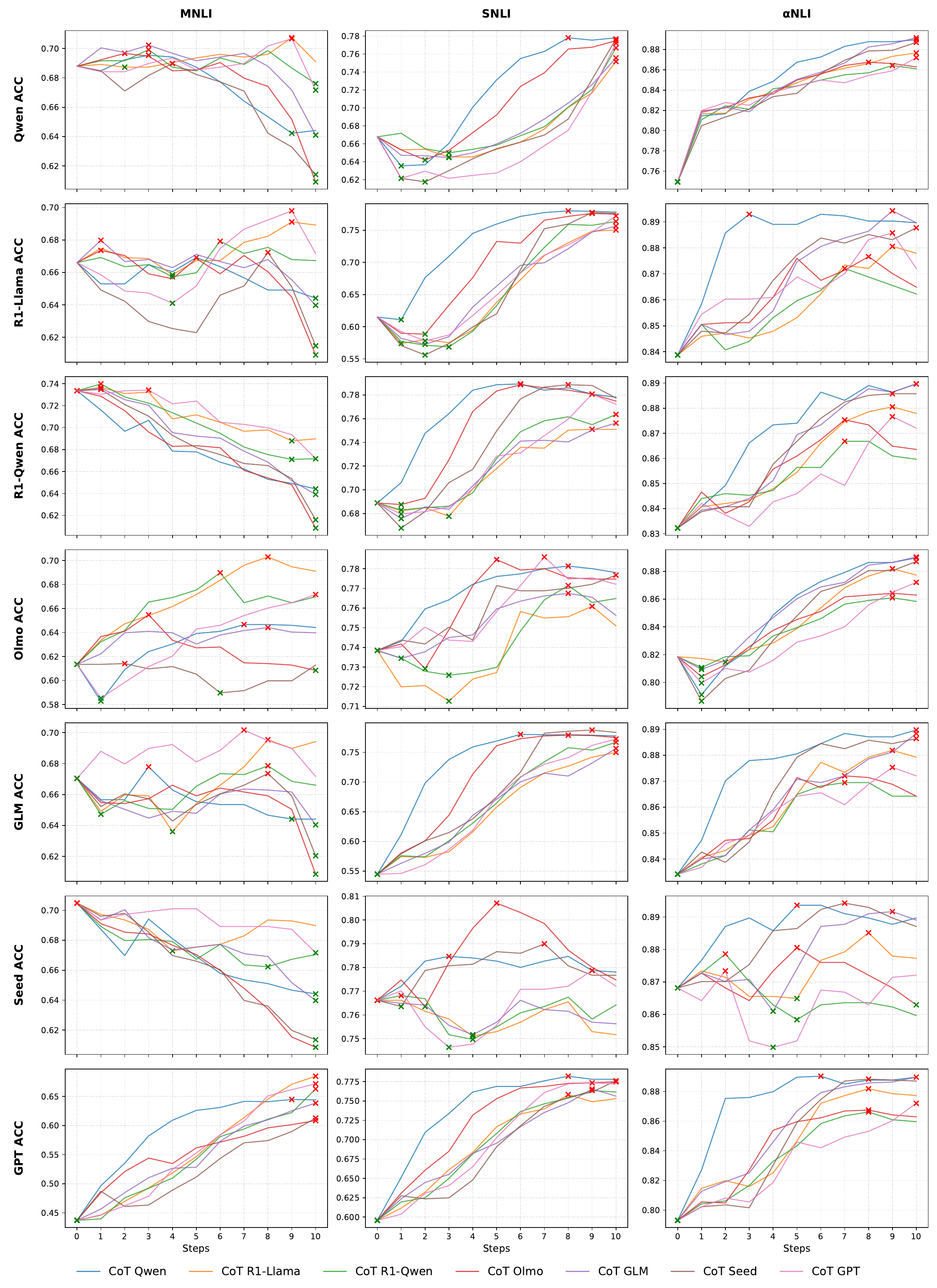}
        \caption{Curves ACC.}
        \label{fig:curve-acc}
\end{figure*}

\begin{figure*}[t]

        \centering
        \includegraphics[width=\linewidth]{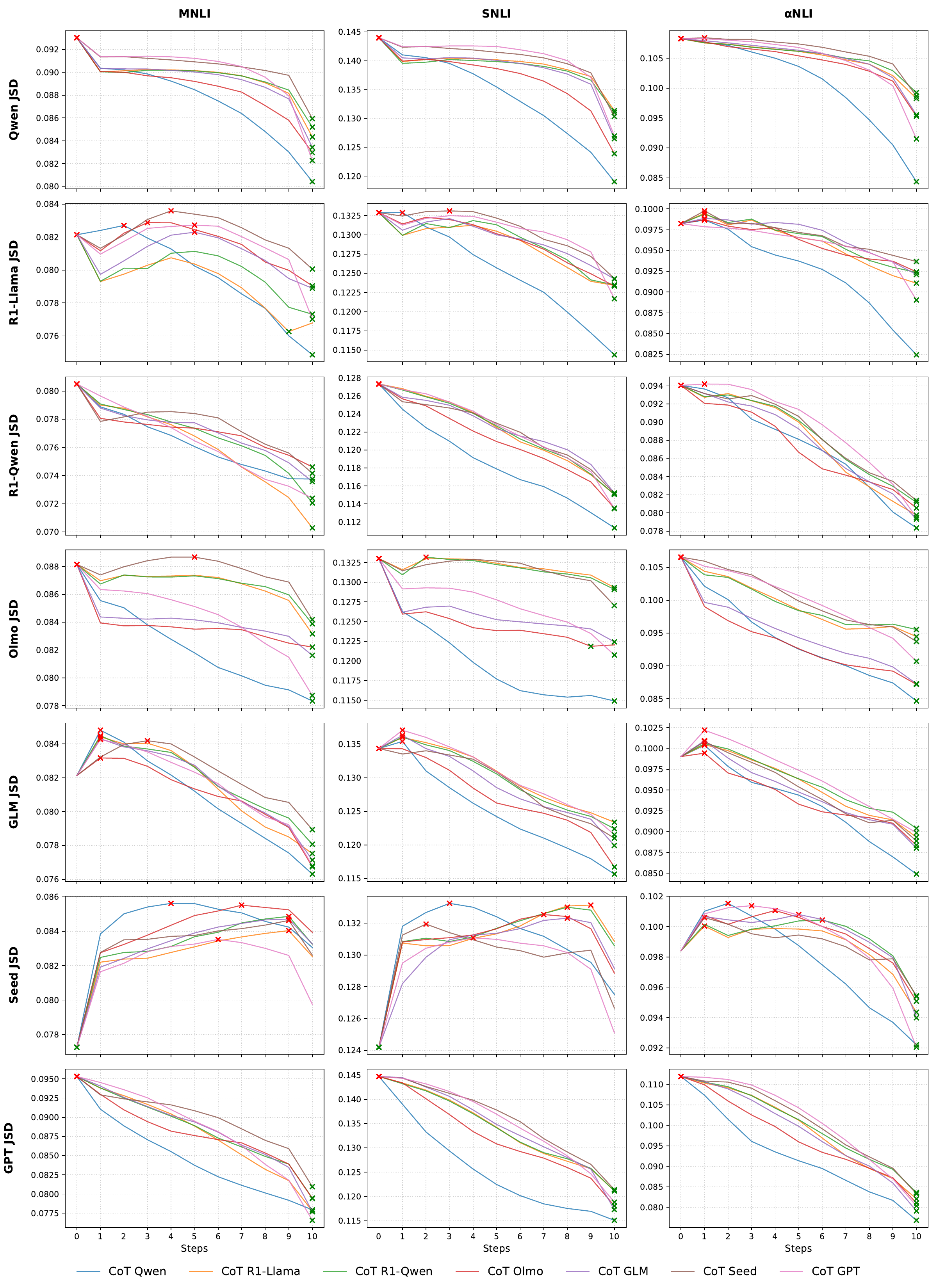}
        \caption{Curves JSD.}
        \label{fig:curve-jsd}
\end{figure*}

\begin{figure*}[t]

        \centering
        \includegraphics[width=\linewidth]{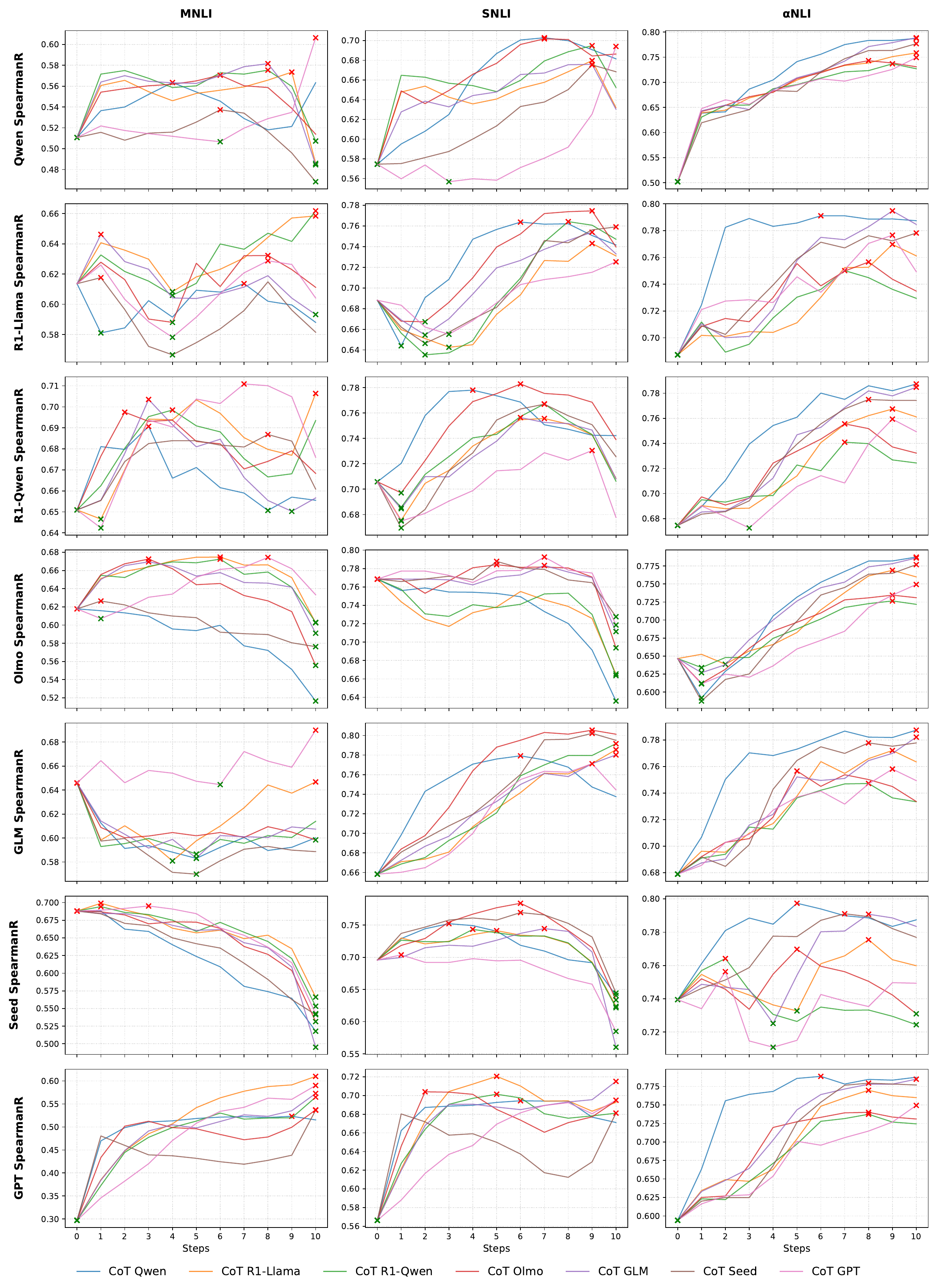}
        \caption{Curves Spearman’s $\rho$.}
        \label{fig:curve-spermanR}
\end{figure*}

This section presents all curves obtained from the step-wise evaluation, providing a complementary perspective to the heatmaps.
Specifically, we show the progression of \textbf{accuracy} (Figure~\ref{fig:curve-acc}), \textbf{JSD}~\ref{fig:curve-jsd}, and \textbf{Spearman’s $\rho$} (Figure~\ref{fig:curve-spermanR}) across inference steps. 
These curves allow us to track how each metric evolves throughout the reasoning process, revealing dynamic trends in model performance, distributional alignment, and rank correlation over time.

\section{Analysis of CoT Formats}
\label{sec:appendix_qualitative}

To further investigate the ``split influence'' observed in our quantitative results---where CoT determines the final accuracy but leaves the distributional structure (JSD and Spearman's $\rho$) largely anchored to model priors---we conducted an examination of the generated reasoning traces.

As discussed in Section~\ref{sec:when}, step-wise analysis reveals that accuracy often shifts or converges only at the conclusion, while Spearman's $\rho$ fluctuates without a consistent trend. Thus, CoT determines the LLM's final choice but not the structure of uncertainty. We attribute this to the structural format of CoT: reasoning traces typically culminate in explicit, decisive conclusion statements (e.g., ``Therefore, the answer is...''), which strongly steer the accuracy in the final steps.

Table \ref{tab:cot_qualitative} presents anecdotal evidence from the $\alpha$NLI dataset. We observe a consistent pattern across models:
\begin{itemize}
    \item \textbf{Explicit Conclusion at the End:} The reasoning process consistently ends with a strong, definitive statement identifying the correct option (e.g., ``So, A must be the correct choice''). This explicit signal aligns with the sharp rise in accuracy observed in the final steps of our step-wise analysis.
    \item \textbf{Implicit Distributional Weighing:} While the models argue \textit{for} the best option, they rarely explicitly articulate the relative probability of the alternative options in a way that would restructure the output distribution. Consequently, while the final decision is explicitly dictated by the CoT's conclusion, the distributional structure over non-argmax options remains latent and implicit, governed largely by the model's intrinsic priors.
\end{itemize}

\begin{table*}[t]
    \centering
    \small
    \resizebox{\linewidth}{!}{
    \renewcommand{\arraystretch}{1.4}
    \begin{tabular}{p{0.20\linewidth} p{0.40\linewidth} p{0.32\linewidth}}
        \toprule
        \textbf{Model \& Case} & \textbf{Reasoning Excerpt (Conclusion Phase)} & \textbf{Observation regarding Final Choice vs. Structure} \\
        \midrule
        \textbf{Qwen} \newline (Instance 2: Sandy) & 
        ``...Therefore, A is better. I think B is a distractor. [...] The instruction says `select ONE of the listed options'... \textbf{So, my response should be just `A'.}'' & 
        \textbf{Decisive Locking:} The reasoning concludes by explicitly discarding the alternative and locking onto the single target token `A', driving the final accuracy without refining the relative probability space. \\
        \midrule
        \textbf{GLM} \newline (Instance 3: Bananas) & 
        ``...Therefore, B cannot explain why they're talking about eating a banana. \textbf{So, A must be the correct choice.} [...] \textbf{Therefore, A is correct.}'' & 
        \textbf{Convergence to Argmax:} The trace culminates in strong assertions (``must be'', ``correct''), which serve to fix the model's final decision, explaining why accuracy converges at the end while latent distributions remain implicit. \\
        \midrule
        \textbf{GPT} \newline (Instance 1: Ron) & 
        ``...That suggests his actions. So A is more appropriate. \textbf{Thus choose A.} [...] We must start answer with a single letter... \textbf{So final answer: `A'.}'' & 
        \textbf{Explicit Selection:} The reasoning shifts from semantic evaluation to an operational selection command (``Thus choose A''), confirming that the CoT acts as a decision-maker for the top option. \\
        \bottomrule
    \end{tabular}}
\caption{Examples of CoT reasoning traces from the $\alpha$NLI dataset. The excerpts illustrate how CoT reasoning typically ends with explicit conclusion statements. This structural characteristic supports our finding that CoT content determines the final choice (Accuracy) through explicit reasoning, while the underlying structure of uncertainty (JSD/Ranking) remains latent and less affected by these definitive concluding remarks.}
    \label{tab:cot_qualitative}
\end{table*}


\begin{thebibliography}{51}
\providecommand{\natexlab}[1]{#1}

\bibitem[{Aroyo and Welty(2015)}]{DBLP:journals/aim/AroyoW15}
Lora Aroyo and Chris Welty. 2015.
\newblock \href {https://doi.org/10.1609/AIMAG.V36I1.2564} {Truth is a lie: Crowd truth and the seven myths of human annotation}.
\newblock \emph{{AI} Mag.}, 36(1):15--24.

\bibitem[{Bhagavatula et~al.(2020)Bhagavatula, Bras, Malaviya, Sakaguchi, Holtzman, Rashkin, Downey, Yih, and Choi}]{DBLP:conf/iclr/BhagavatulaBMSH20}
Chandra Bhagavatula, Ronan~Le Bras, Chaitanya Malaviya, Keisuke Sakaguchi, Ari Holtzman, Hannah Rashkin, Doug Downey, Wen{-}tau Yih, and Yejin Choi. 2020.
\newblock \href {https://openreview.net/forum?id=Byg1v1HKDB} {Abductive commonsense reasoning}.
\newblock In \emph{8th International Conference on Learning Representations, {ICLR} 2020, Addis Ababa, Ethiopia, April 26-30, 2020}. OpenReview.net.

\bibitem[{Bowman et~al.(2015)Bowman, Angeli, Potts, and Manning}]{bowman-etal-2015-large}
Samuel~R. Bowman, Gabor Angeli, Christopher Potts, and Christopher~D. Manning. 2015.
\newblock \href {https://doi.org/10.18653/v1/D15-1075} {A large annotated corpus for learning natural language inference}.
\newblock In \emph{Proceedings of the 2015 Conference on Empirical Methods in Natural Language Processing}, pages 632--642, Lisbon, Portugal. Association for Computational Linguistics.

\bibitem[{Cabitza et~al.(2023)Cabitza, Campagner, and Basile}]{cabitza2023toward}
Federico Cabitza, Andrea Campagner, and Valerio Basile. 2023.
\newblock \href {https://doi.org/10.1609/AAAI.V37I6.25840} {Toward a perspectivist turn in ground truthing for predictive computing}.
\newblock In \emph{Thirty-Seventh {AAAI} Conference on Artificial Intelligence, {AAAI} 2023, Thirty-Fifth Conference on Innovative Applications of Artificial Intelligence, {IAAI} 2023, Thirteenth Symposium on Educational Advances in Artificial Intelligence, {EAAI} 2023, Washington, DC, USA, February 7-14, 2023}, pages 6860--6868. {AAAI} Press.

\bibitem[{Chen et~al.(2025{\natexlab{a}})Chen, Liu, Korhonen, and Plank}]{chen-etal-2025-threading}
Beiduo Chen, Yang~Janet Liu, Anna Korhonen, and Barbara Plank. 2025{\natexlab{a}}.
\newblock \href {https://doi.org/10.18653/v1/2025.emnlp-main.1682} {Threading the needle: Reweaving chain-of-thought reasoning to explain human label variation}.
\newblock In \emph{Proceedings of the 2025 Conference on Empirical Methods in Natural Language Processing}, pages 33099--33123, Suzhou, China. Association for Computational Linguistics.

\bibitem[{Chen et~al.(2025{\natexlab{b}})Chen, Peng, Korhonen, and Plank}]{chen-etal-2025-rose}
Beiduo Chen, Siyao Peng, Anna Korhonen, and Barbara Plank. 2025{\natexlab{b}}.
\newblock \href {https://doi.org/10.18653/v1/2025.findings-acl.562} {A rose by any other name: {LLM}-generated explanations are good proxies for human explanations to collect label distributions on {NLI}}.
\newblock In \emph{Findings of the Association for Computational Linguistics: ACL 2025}, pages 10777--10802, Vienna, Austria. Association for Computational Linguistics.

\bibitem[{Chen et~al.(2024)Chen, Wang, Peng, Litschko, Korhonen, and Plank}]{chen-etal-2024-seeing}
Beiduo Chen, Xinpeng Wang, Siyao Peng, Robert Litschko, Anna Korhonen, and Barbara Plank. 2024.
\newblock \href {https://doi.org/10.18653/v1/2024.findings-emnlp.842} {``seeing the big through the small'': Can {LLM}s approximate human judgment distributions on {NLI} from a few explanations?}
\newblock In \emph{Findings of the Association for Computational Linguistics: EMNLP 2024}, pages 14396--14419, Miami, Florida, USA. Association for Computational Linguistics.

\bibitem[{DeepSeek{-}AI et~al.(2025)DeepSeek{-}AI, Guo, Yang, Zhang, Song, Zhang, Xu, Zhu, Ma, Wang, Bi, Zhang, Yu, Wu, Wu, Gou, Shao, Li, Gao, Liu, Xue, Wang, Wu, Feng, Lu, Zhao, Deng, Zhang, Ruan, Dai, Chen, Ji, Li, Lin, Dai, Luo, Hao, Chen, Li, Zhang, Bao, Xu, Wang, Ding, Xin, Gao, Qu, Li, Guo, Li, Wang, Chen, Yuan, Qiu, Li, Cai, Ni, Liang, Chen, Dong, Hu, Gao, Guan, Huang, Yu, Wang, Zhang, Zhao, Wang, Zhang, Xu, Xia, Zhang, Zhang, Tang, Li, Wang, Li, Tian, Huang, Zhang, Wang, Chen, Du, Ge, Zhang, Pan, Wang, Chen, Jin, Chen, Lu, Zhou, Chen, Ye, Wang, Yu, Zhou, Pan, and Li}]{DBLP:journals/corr/abs-2501-12948}
DeepSeek{-}AI, Daya Guo, Dejian Yang, Haowei Zhang, Junxiao Song, Ruoyu Zhang, Runxin Xu, Qihao Zhu, Shirong Ma, Peiyi Wang, Xiao Bi, Xiaokang Zhang, Xingkai Yu, Yu~Wu, Z.~F. Wu, Zhibin Gou, Zhihong Shao, Zhuoshu Li, Ziyi Gao, and 81 others. 2025.
\newblock \href {https://doi.org/10.48550/ARXIV.2501.12948} {Deepseek-r1: Incentivizing reasoning capability in llms via reinforcement learning}.
\newblock \emph{CoRR}, abs/2501.12948.

\bibitem[{Dong et~al.(2026)Dong, Hu, Hui, Zhang, Vulic, Bobu, and Collier}]{DBLP:journals/corr/abs-2601-06407}
Yijiang~River Dong, Tiancheng Hu, Zheng Hui, Caiqi Zhang, Ivan Vulic, Andreea Bobu, and Nigel Collier. 2026.
\newblock \href {https://doi.org/10.48550/ARXIV.2601.06407} {Value of information: {A} framework for human-agent communication}.
\newblock \emph{CoRR}, abs/2601.06407.

\bibitem[{Dubey et~al.(2024)Dubey, Jauhri, Pandey, Kadian, Al{-}Dahle, Letman, Mathur, Schelten, Yang, Fan, Goyal, Hartshorn, Yang, Mitra, Sravankumar, Korenev, Hinsvark, Rao, Zhang, Rodriguez, Gregerson, Spataru, Rozi{\`{e}}re, Biron, Tang, Chern, Caucheteux, Nayak, Bi, Marra, McConnell, Keller, Touret, Wu, Wong, Ferrer, Nikolaidis, Allonsius, Song, Pintz, Livshits, Esiobu, Choudhary, Mahajan, Garcia{-}Olano, Perino, Hupkes, Lakomkin, AlBadawy, Lobanova, Dinan, Smith, Radenovic, Zhang, Synnaeve, Lee, Anderson, Nail, Mialon, Pang, Cucurell, Nguyen, Korevaar, Xu, Touvron, Zarov, Ibarra, Kloumann, Misra, Evtimov, Copet, Lee, Geffert, Vranes, Park, Mahadeokar, Shah, van~der Linde, Billock, Hong, Lee, Fu, Chi, Huang, Liu, Wang, Yu, Bitton, Spisak, Park, Rocca, Johnstun, Saxe, Jia, Alwala, Upasani, Plawiak, Li, Heafield, Stone, and et~al.}]{DBLP:journals/corr/abs-2407-21783}
Abhimanyu Dubey, Abhinav Jauhri, Abhinav Pandey, Abhishek Kadian, Ahmad Al{-}Dahle, Aiesha Letman, Akhil Mathur, Alan Schelten, Amy Yang, Angela Fan, Anirudh Goyal, Anthony Hartshorn, Aobo Yang, Archi Mitra, Archie Sravankumar, Artem Korenev, Arthur Hinsvark, Arun Rao, Aston Zhang, and 82 others. 2024.
\newblock \href {https://doi.org/10.48550/ARXIV.2407.21783} {{The Llama 3 Herd of Models}}.
\newblock \emph{CoRR}, abs/2407.21783.

\bibitem[{Durmus et~al.(2023)Durmus, Nyugen, Liao, Schiefer, Askell, andAbhimanyu Carol~Chen, Hatfield{-}Dodds, Hernandez, Joseph, Lovitt, McCandlish, Sikder, Tamkin, Thamkul, Kaplan, Clark, and Ganguli}]{DBLP:journals/corr/abs-2306-16388}
Esin Durmus, Karina Nyugen, Thomas~I. Liao, Nicholas Schiefer, Amanda Askell, Anton~Bakhtin andAbhimanyu Carol~Chen, Zac Hatfield{-}Dodds, Danny Hernandez, Nicholas Joseph, Liane Lovitt, Sam McCandlish, Orowa Sikder, Alex Tamkin, Janel Thamkul, Jared Kaplan, Jack Clark, and Deep Ganguli. 2023.
\newblock \href {https://doi.org/10.48550/ARXIV.2306.16388} {Towards measuring the representation of subjective global opinions in language models}.
\newblock \emph{CoRR}, abs/2306.16388.

\bibitem[{Endres and Schindelin(2003)}]{DBLP:journals/tit/EndresS03}
Dominik~Maria Endres and Johannes~E. Schindelin. 2003.
\newblock \href {https://doi.org/10.1109/TIT.2003.813506} {A new metric for probability distributions}.
\newblock \emph{{IEEE} Trans. Inf. Theory}, 49(7):1858--1860.

\bibitem[{GLM et~al.(2024)GLM, Zeng, Xu, Wang, Zhang, Yin, Rojas, Feng, Zhao, Lai, Yu, Wang, Sun, Zhang, Cheng, Gui, Tang, Zhang, Li, Zhao, Wu, Zhong, Liu, Huang, Zhang, Zheng, Lu, Duan, Zhang, Cao, Yang, Tam, Zhao, Liu, Xia, Zhang, Gu, Lv, Liu, Liu, Yang, Song, Zhang, An, Xu, Niu, Yang, Li, Bai, Dong, Qi, Wang, Yang, Du, Hou, and Wang}]{glm2024chatglm}
Team GLM, Aohan Zeng, Bin Xu, Bowen Wang, Chenhui Zhang, Da~Yin, Diego Rojas, Guanyu Feng, Hanlin Zhao, Hanyu Lai, Hao Yu, Hongning Wang, Jiadai Sun, Jiajie Zhang, Jiale Cheng, Jiayi Gui, Jie Tang, Jing Zhang, Juanzi Li, and 37 others. 2024.
\newblock \href {https://arxiv.org/abs/2406.12793} {Chatglm: A family of large language models from glm-130b to glm-4 all tools}.
\newblock \emph{Preprint}, arXiv:2406.12793.

\bibitem[{Hendrycks et~al.(2021{\natexlab{a}})Hendrycks, Burns, Basart, Zou, Mazeika, Song, and Steinhardt}]{DBLP:conf/iclr/HendrycksBBZMSS21}
Dan Hendrycks, Collin Burns, Steven Basart, Andy Zou, Mantas Mazeika, Dawn Song, and Jacob Steinhardt. 2021{\natexlab{a}}.
\newblock \href {https://openreview.net/forum?id=d7KBjmI3GmQ} {Measuring massive multitask language understanding}.
\newblock In \emph{9th International Conference on Learning Representations, {ICLR} 2021, Virtual Event, Austria, May 3-7, 2021}. OpenReview.net.

\bibitem[{Hendrycks et~al.(2021{\natexlab{b}})Hendrycks, Burns, Kadavath, Arora, Basart, Tang, Song, and Steinhardt}]{DBLP:conf/nips/HendrycksBKABTS21}
Dan Hendrycks, Collin Burns, Saurav Kadavath, Akul Arora, Steven Basart, Eric Tang, Dawn Song, and Jacob Steinhardt. 2021{\natexlab{b}}.
\newblock \href {https://datasets-benchmarks-proceedings.neurips.cc/paper/2021/hash/be83ab3ecd0db773eb2dc1b0a17836a1-Abstract-round2.html} {Measuring mathematical problem solving with the {MATH} dataset}.
\newblock In \emph{Proceedings of the Neural Information Processing Systems Track on Datasets and Benchmarks 1, NeurIPS Datasets and Benchmarks 2021, December 2021, virtual}.

\bibitem[{Hong et~al.(2025{\natexlab{a}})Hong, Chen, Peng, de~Marneffe, and Plank}]{hong-etal-2025-litex}
Pingjun Hong, Beiduo Chen, Siyao Peng, Marie-Catherine de~Marneffe, and Barbara Plank. 2025{\natexlab{a}}.
\newblock \href {https://doi.org/10.18653/v1/2025.emnlp-main.1728} {{L}i{TE}x: A linguistic taxonomy of explanations for understanding within-label variation in natural language inference}.
\newblock In \emph{Proceedings of the 2025 Conference on Empirical Methods in Natural Language Processing}, pages 34053--34073, Suzhou, China. Association for Computational Linguistics.

\bibitem[{Hong et~al.(2025{\natexlab{b}})Hong, Chen, Peng, de~Marneffe, Roth, and Plank}]{DBLP:journals/corr/abs-2510-16458}
Pingjun Hong, Beiduo Chen, Siyao Peng, Marie{-}Catherine de~Marneffe, Benjamin Roth, and Barbara Plank. 2025{\natexlab{b}}.
\newblock \href {https://doi.org/10.48550/ARXIV.2510.16458} {Agree, disagree, explain: Decomposing human label variation in {NLI} through the lens of explanations}.
\newblock \emph{CoRR}, abs/2510.16458.

\bibitem[{Hu et~al.(2025)Hu, Baumann, Lupo, Collier, Hovy, and R{\"{o}}ttger}]{DBLP:journals/corr/abs-2510-17516}
Tiancheng Hu, Joachim Baumann, Lorenzo Lupo, Nigel Collier, Dirk Hovy, and Paul R{\"{o}}ttger. 2025.
\newblock \href {https://doi.org/10.48550/ARXIV.2510.17516} {Simbench: Benchmarking the ability of large language models to simulate human behaviors}.
\newblock \emph{CoRR}, abs/2510.17516.

\bibitem[{Hurst et~al.(2024)Hurst, Lerer, Goucher, Perelman, Ramesh, Clark, Ostrow, Welihinda, Hayes, Radford, Madry, Baker{-}Whitcomb, Beutel, Borzunov, Carney, Chow, Kirillov, Nichol, Paino, Renzin, Passos, Kirillov, Christakis, Conneau, Kamali, Jabri, Moyer, Tam, Crookes, Tootoonchian, Kumar, Vallone, Karpathy, Braunstein, Cann, Codispoti, Galu, Kondrich, Tulloch, Mishchenko, Baek, Jiang, Pelisse, Woodford, Gosalia, Dhar, Pantuliano, Nayak, Oliver, Zoph, Ghorbani, Leimberger, Rossen, Sokolowsky, Wang, Zweig, Hoover, Samic, McGrew, Spero, Giertler, Cheng, Lightcap, Walkin, Quinn, Guarraci, Hsu, Kellogg, Eastman, Lugaresi, Wainwright, Bassin, Hudson, Chu, Nelson, Li, Shern, Conger, Barette, Voss, Ding, Lu, Zhang, Beaumont, Hallacy, Koch, Gibson, Kim, Choi, McLeavey, Hesse, Fischer, Winter, Czarnecki, Jarvis, Wei, Koumouzelis, and Sherburn}]{DBLP:journals/corr/abs-2410-21276}
Aaron Hurst, Adam Lerer, Adam~P. Goucher, Adam Perelman, Aditya Ramesh, Aidan Clark, AJ~Ostrow, Akila Welihinda, Alan Hayes, Alec Radford, Aleksander Madry, Alex Baker{-}Whitcomb, Alex Beutel, Alex Borzunov, Alex Carney, Alex Chow, Alex Kirillov, Alex Nichol, Alex Paino, and 79 others. 2024.
\newblock \href {https://doi.org/10.48550/ARXIV.2410.21276} {{GPT-4o System Card}}.
\newblock \emph{CoRR}, abs/2410.21276.

\bibitem[{Jiang et~al.(2023)Jiang, Tan, and de~Marneffe}]{jiang-etal-2023-ecologically}
Nan-Jiang Jiang, Chenhao Tan, and Marie-Catherine de~Marneffe. 2023.
\newblock \href {https://doi.org/10.18653/v1/2023.findings-emnlp.712} {Ecologically valid explanations for label variation in {NLI}}.
\newblock In \emph{Findings of the Association for Computational Linguistics: EMNLP 2023}, pages 10622--10633, Singapore. Association for Computational Linguistics.

\bibitem[{Kurniawan et~al.(2025)Kurniawan, Mistica, Baldwin, and Lau}]{DBLP:journals/corr/abs-2502-01891}
Kemal Kurniawan, Meladel Mistica, Timothy Baldwin, and Jey~Han Lau. 2025.
\newblock \href {https://doi.org/10.48550/ARXIV.2502.01891} {Training and evaluating with human label variation: An empirical study}.
\newblock \emph{CoRR}, abs/2502.01891.

\bibitem[{Lee et~al.(2023)Lee, An, and Thorne}]{lee-etal-2023-large}
Noah Lee, Na~Min An, and James Thorne. 2023.
\newblock \href {https://doi.org/10.18653/v1/2023.emnlp-main.278} {Can large language models capture dissenting human voices?}
\newblock In \emph{Proceedings of the 2023 Conference on Empirical Methods in Natural Language Processing}, pages 4569--4585, Singapore. Association for Computational Linguistics.

\bibitem[{Leonardelli et~al.(2023)Leonardelli, Abercrombie, Almanea, Basile, Fornaciari, Plank, Rieser, Uma, and Poesio}]{leonardelli-etal-2023-semeval}
Elisa Leonardelli, Gavin Abercrombie, Dina Almanea, Valerio Basile, Tommaso Fornaciari, Barbara Plank, Verena Rieser, Alexandra Uma, and Massimo Poesio. 2023.
\newblock \href {https://doi.org/10.18653/v1/2023.semeval-1.314} {{S}em{E}val-2023 task 11: Learning with disagreements ({L}e{W}i{D}i)}.
\newblock In \emph{Proceedings of the 17th International Workshop on Semantic Evaluation (SemEval-2023)}, pages 2304--2318, Toronto, Canada. Association for Computational Linguistics.

\bibitem[{Liang et~al.(2023)Liang, Bommasani, Lee, Tsipras, Soylu, Yasunaga, Zhang, Narayanan, Wu, Kumar, Newman, Yuan, Yan, Zhang, Cosgrove, Manning, R{\'{e}}, Acosta{-}Navas, Hudson, Zelikman, Durmus, Ladhak, Rong, Ren, Yao, Wang, Santhanam, Orr, Zheng, Y{\"{u}}ksekg{\"{o}}n{\"{u}}l, Suzgun, Kim, Guha, Chatterji, Khattab, Henderson, Huang, Chi, Xie, Santurkar, Ganguli, Hashimoto, Icard, Zhang, Chaudhary, Wang, Li, Mai, Zhang, and Koreeda}]{DBLP:journals/tmlr/LiangBLTSYZNWKN23}
Percy Liang, Rishi Bommasani, Tony Lee, Dimitris Tsipras, Dilara Soylu, Michihiro Yasunaga, Yian Zhang, Deepak Narayanan, Yuhuai Wu, Ananya Kumar, Benjamin Newman, Binhang Yuan, Bobby Yan, Ce~Zhang, Christian Cosgrove, Christopher~D. Manning, Christopher R{\'{e}}, Diana Acosta{-}Navas, Drew~A. Hudson, and 31 others. 2023.
\newblock \href {https://openreview.net/forum?id=iO4LZibEqW} {Holistic evaluation of language models}.
\newblock \emph{Trans. Mach. Learn. Res.}, 2023.

\bibitem[{Liu et~al.(2026)Liu, Zhao, Sch{\"{u}}tze, and Hedderich}]{DBLP:journals/corr/abs-2601-02996}
Yihong Liu, Raoyuan Zhao, Hinrich Sch{\"{u}}tze, and Michael~A. Hedderich. 2026.
\newblock \href {https://doi.org/10.48550/ARXIV.2601.02996} {Large reasoning models are (not yet) multilingual latent reasoners}.
\newblock \emph{CoRR}, abs/2601.02996.

\bibitem[{Mao et~al.(2025{\natexlab{a}})Mao, Yin, Zhu, and Fang}]{DBLP:journals/corr/abs-2509-14004}
Minjia Mao, Bowen Yin, Yu~Zhu, and Xiao Fang. 2025{\natexlab{a}}.
\newblock \href {https://doi.org/10.48550/ARXIV.2509.14004} {Early stopping chain-of-thoughts in large language models}.
\newblock \emph{CoRR}, abs/2509.14004.

\bibitem[{Mao et~al.(2025{\natexlab{b}})Mao, Bisliouk, Nama, and Ruchkin}]{DBLP:journals/corr/abs-2506-08243}
Zhenjiang Mao, Artem Bisliouk, Rohith~Reddy Nama, and Ivan Ruchkin. 2025{\natexlab{b}}.
\newblock \href {https://doi.org/10.48550/ARXIV.2506.08243} {Temporalizing confidence: Evaluation of chain-of-thought reasoning with signal temporal logic}.
\newblock \emph{CoRR}, abs/2506.08243.

\bibitem[{Ni et~al.(2025)Ni, Fan, Zouhar, Rooein, Hoyle, Sachan, Leippold, Hovy, and Ash}]{ni2025can}
Jingwei Ni, Yu~Fan, Vilém Zouhar, Donya Rooein, Alexander Hoyle, Mrinmaya Sachan, Markus Leippold, Dirk Hovy, and Elliott Ash. 2025.
\newblock \href {https://arxiv.org/abs/2506.19467} {Can reasoning help large language models capture human annotator disagreement?}
\newblock \emph{Preprint}, arXiv:2506.19467.

\bibitem[{Nie et~al.(2020)Nie, Zhou, and Bansal}]{nie-etal-2020-learn}
Yixin Nie, Xiang Zhou, and Mohit Bansal. 2020.
\newblock \href {https://doi.org/10.18653/v1/2020.emnlp-main.734} {What can we learn from collective human opinions on natural language inference data?}
\newblock In \emph{Proceedings of the 2020 Conference on Empirical Methods in Natural Language Processing (EMNLP)}, pages 9131--9143, Online. Association for Computational Linguistics.

\bibitem[{Olmo et~al.(2025)Olmo, Ettinger, Bertsch, Kuehl, Graham, Heineman, Groeneveld, Brahman, Timbers, Ivison et~al.}]{olmo2025olmo}
Team Olmo, Allyson Ettinger, Amanda Bertsch, Bailey Kuehl, David Graham, David Heineman, Dirk Groeneveld, Faeze Brahman, Finbarr Timbers, Hamish Ivison, and 1 others. 2025.
\newblock Olmo 3.
\newblock \emph{arXiv preprint arXiv:2512.13961}.

\bibitem[{OpenAI(2023)}]{DBLP:journals/corr/abs-2303-08774}
OpenAI. 2023.
\newblock \href {https://doi.org/10.48550/ARXIV.2303.08774} {{GPT-4} technical report}.
\newblock \emph{CoRR}, abs/2303.08774.

\bibitem[{OpenAI(2025)}]{openai2025gptoss120bgptoss20bmodel}
OpenAI. 2025.
\newblock \href {https://arxiv.org/abs/2508.10925} {gpt-oss-120b \& gpt-oss-20b model card}.
\newblock \emph{Preprint}, arXiv:2508.10925.

\bibitem[{Pavlick and Kwiatkowski(2019)}]{pavlick-kwiatkowski-2019-inherent}
Ellie Pavlick and Tom Kwiatkowski. 2019.
\newblock \href {https://doi.org/10.1162/tacl_a_00293} {Inherent disagreements in human textual inferences}.
\newblock \emph{Transactions of the Association for Computational Linguistics}, 7:677--694.

\bibitem[{Plank(2022)}]{plank-2022-problem}
Barbara Plank. 2022.
\newblock \href {https://doi.org/10.18653/v1/2022.emnlp-main.731} {The ``problem'' of human label variation: On ground truth in data, modeling and evaluation}.
\newblock In \emph{Proceedings of the 2022 Conference on Empirical Methods in Natural Language Processing}, pages 10671--10682, Abu Dhabi, United Arab Emirates. Association for Computational Linguistics.

\bibitem[{Rein et~al.(2023)Rein, Hou, Stickland, Petty, Pang, Dirani, Michael, and Bowman}]{DBLP:journals/corr/abs-2311-12022}
David Rein, Betty~Li Hou, Asa~Cooper Stickland, Jackson Petty, Richard~Yuanzhe Pang, Julien Dirani, Julian Michael, and Samuel~R. Bowman. 2023.
\newblock \href {https://doi.org/10.48550/ARXIV.2311.12022} {{GPQA:} {A} graduate-level google-proof q{\&}a benchmark}.
\newblock \emph{CoRR}, abs/2311.12022.

\bibitem[{Santurkar et~al.(2023)Santurkar, Durmus, Ladhak, Lee, Liang, and Hashimoto}]{DBLP:conf/icml/SanturkarDLLLH23}
Shibani Santurkar, Esin Durmus, Faisal Ladhak, Cinoo Lee, Percy Liang, and Tatsunori Hashimoto. 2023.
\newblock \href {https://proceedings.mlr.press/v202/santurkar23a.html} {Whose opinions do language models reflect?}
\newblock In \emph{International Conference on Machine Learning, {ICML} 2023, 23-29 July 2023, Honolulu, Hawaii, {USA}}, volume 202 of \emph{Proceedings of Machine Learning Research}, pages 29971--30004. {PMLR}.

\bibitem[{Spearman(1961)}]{spearman1961proof}
Charles Spearman. 1961.
\newblock The proof and measurement of association between two things.

\bibitem[{Sun et~al.(2025)Sun, Min, Chen, Zhao, Liu, Wang, Fang, and Wen}]{DBLP:journals/corr/abs-2503-21380}
Haoxiang Sun, Yingqian Min, Zhipeng Chen, Wayne~Xin Zhao, Zheng Liu, Zhongyuan Wang, Lei Fang, and Ji{-}Rong Wen. 2025.
\newblock \href {https://doi.org/10.48550/ARXIV.2503.21380} {Challenging the boundaries of reasoning: An olympiad-level math benchmark for large language models}.
\newblock \emph{CoRR}, abs/2503.21380.

\bibitem[{Team(2025{\natexlab{a}})}]{seed2025seed-oss}
ByteDance~Seed Team. 2025{\natexlab{a}}.
\newblock Seed-oss open-source models.
\newblock \url{https://github.com/ByteDance-Seed/seed-oss}.

\bibitem[{Team(2025{\natexlab{b}})}]{qwen3technicalreport}
Qwen Team. 2025{\natexlab{b}}.
\newblock \href {https://arxiv.org/abs/2505.09388} {Qwen3 technical report}.
\newblock \emph{Preprint}, arXiv:2505.09388.

\bibitem[{Team(2025{\natexlab{c}})}]{qwq32b}
Qwen Team. 2025{\natexlab{c}}.
\newblock \href {https://qwenlm.github.io/blog/qwq-32b/} {Qwq-32b: Embracing the power of reinforcement learning}.

\bibitem[{Touvron et~al.(2023)Touvron, Martin, Stone, Albert, Almahairi, Babaei, Bashlykov, Batra, Bhargava, Bhosale, Bikel, Blecher, Canton{-}Ferrer, Chen, Cucurull, Esiobu, Fernandes, Fu, Fu, Fuller, Gao, Goswami, Goyal, Hartshorn, Hosseini, Hou, Inan, Kardas, Kerkez, Khabsa, Kloumann, Korenev, Koura, Lachaux, Lavril, Lee, Liskovich, Lu, Mao, Martinet, Mihaylov, Mishra, Molybog, Nie, Poulton, Reizenstein, Rungta, Saladi, Schelten, Silva, Smith, Subramanian, Tan, Tang, Taylor, Williams, Kuan, Xu, Yan, Zarov, Zhang, Fan, Kambadur, Narang, Rodriguez, Stojnic, Edunov, and Scialom}]{DBLP:journals/corr/abs-2307-09288}
Hugo Touvron, Louis Martin, Kevin Stone, Peter Albert, Amjad Almahairi, Yasmine Babaei, Nikolay Bashlykov, Soumya Batra, Prajjwal Bhargava, Shruti Bhosale, Dan Bikel, Lukas Blecher, Cristian Canton{-}Ferrer, Moya Chen, Guillem Cucurull, David Esiobu, Jude Fernandes, Jeremy Fu, Wenyin Fu, and 49 others. 2023.
\newblock \href {https://doi.org/10.48550/ARXIV.2307.09288} {Llama 2: Open foundation and fine-tuned chat models}.
\newblock \emph{CoRR}, abs/2307.09288.

\bibitem[{Uma et~al.(2021)Uma, Fornaciari, Hovy, Paun, Plank, and Poesio}]{DBLP:journals/jair/UmaFHPPP21}
Alexandra Uma, Tommaso Fornaciari, Dirk Hovy, Silviu Paun, Barbara Plank, and Massimo Poesio. 2021.
\newblock \href {https://doi.org/10.1613/JAIR.1.12752} {Learning from disagreement: {A} survey}.
\newblock \emph{J. Artif. Intell. Res.}, 72:1385--1470.

\bibitem[{Wang et~al.(2023)Wang, Wei, Schuurmans, Le, Chi, Narang, Chowdhery, and Zhou}]{DBLP:conf/iclr/0002WSLCNCZ23}
Xuezhi Wang, Jason Wei, Dale Schuurmans, Quoc~V. Le, Ed~H. Chi, Sharan Narang, Aakanksha Chowdhery, and Denny Zhou. 2023.
\newblock \href {https://openreview.net/forum?id=1PL1NIMMrw} {Self-consistency improves chain of thought reasoning in language models}.
\newblock In \emph{The Eleventh International Conference on Learning Representations, {ICLR} 2023, Kigali, Rwanda, May 1-5, 2023}. OpenReview.net.

\bibitem[{Wang et~al.(2024)Wang, Ma, Zhang, Ni, Chandra, Guo, Ren, Arulraj, He, Jiang, Li, Ku, Wang, Zhuang, Fan, Yue, and Chen}]{DBLP:conf/nips/WangMZNCGRAHJLK24}
Yubo Wang, Xueguang Ma, Ge~Zhang, Yuansheng Ni, Abhranil Chandra, Shiguang Guo, Weiming Ren, Aaran Arulraj, Xuan He, Ziyan Jiang, Tianle Li, Max Ku, Kai Wang, Alex Zhuang, Rongqi Fan, Xiang Yue, and Wenhu Chen. 2024.
\newblock \href {http://papers.nips.cc/paper\_files/paper/2024/hash/ad236edc564f3e3156e1b2feafb99a24-Abstract-Datasets\_and\_Benchmarks\_Track.html} {Mmlu-pro: {A} more robust and challenging multi-task language understanding benchmark}.
\newblock In \emph{Advances in Neural Information Processing Systems 38: Annual Conference on Neural Information Processing Systems 2024, NeurIPS 2024, Vancouver, BC, Canada, December 10 - 15, 2024}.

\bibitem[{Weber-Genzel et~al.(2024)Weber-Genzel, Peng, De~Marneffe, and Plank}]{weber-genzel-etal-2024-varierr}
Leon Weber-Genzel, Siyao Peng, Marie-Catherine De~Marneffe, and Barbara Plank. 2024.
\newblock \href {https://doi.org/10.18653/v1/2024.acl-long.123} {{V}ari{E}rr {NLI}: Separating annotation error from human label variation}.
\newblock In \emph{Proceedings of the 62nd Annual Meeting of the Association for Computational Linguistics (Volume 1: Long Papers)}, pages 2256--2269, Bangkok, Thailand. Association for Computational Linguistics.

\bibitem[{Wei et~al.(2022)Wei, Wang, Schuurmans, Bosma, Ichter, Xia, Chi, Le, and Zhou}]{DBLP:conf/nips/Wei0SBIXCLZ22}
Jason Wei, Xuezhi Wang, Dale Schuurmans, Maarten Bosma, Brian Ichter, Fei Xia, Ed~H. Chi, Quoc~V. Le, and Denny Zhou. 2022.
\newblock \href {http://papers.nips.cc/paper\_files/paper/2022/hash/9d5609613524ecf4f15af0f7b31abca4-Abstract-Conference.html} {Chain-of-thought prompting elicits reasoning in large language models}.
\newblock In \emph{Advances in Neural Information Processing Systems 35: Annual Conference on Neural Information Processing Systems 2022, NeurIPS 2022, New Orleans, LA, USA, November 28 - December 9, 2022}.

\bibitem[{Williams et~al.(2018)Williams, Nangia, and Bowman}]{williams-etal-2018-broad}
Adina Williams, Nikita Nangia, and Samuel Bowman. 2018.
\newblock \href {https://doi.org/10.18653/v1/N18-1101} {A broad-coverage challenge corpus for sentence understanding through inference}.
\newblock In \emph{Proceedings of the 2018 Conference of the North {A}merican Chapter of the Association for Computational Linguistics: Human Language Technologies, Volume 1 (Long Papers)}, pages 1112--1122, New Orleans, Louisiana. Association for Computational Linguistics.

\bibitem[{Yoon et~al.(2025)Yoon, Kim, Yang, Kim, Kim, Kim, Choi, Kim, and Seo}]{DBLP:journals/corr/abs-2505-14489}
Dongkeun Yoon, Seungone Kim, Sohee Yang, Sunkyoung Kim, Soyeon Kim, Yongil Kim, Eunbi Choi, Yireun Kim, and Minjoon Seo. 2025.
\newblock \href {https://doi.org/10.48550/ARXIV.2505.14489} {Reasoning models better express their confidence}.
\newblock \emph{CoRR}, abs/2505.14489.

\bibitem[{Zhao et~al.(2025)Zhao, Tan, Ma, Li, Jiang, Wang, Yang, and Liu}]{DBLP:journals/corr/abs-2508-01191}
Chengshuai Zhao, Zhen Tan, Pingchuan Ma, Dawei Li, Bohan Jiang, Yancheng Wang, Yingzhen Yang, and Huan Liu. 2025.
\newblock \href {https://doi.org/10.48550/ARXIV.2508.01191} {Is chain-of-thought reasoning of llms a mirage? {A} data distribution lens}.
\newblock \emph{CoRR}, abs/2508.01191.

\bibitem[{Zhao et~al.(2026)Zhao, Liu, Schuetze, and Hedderich}]{zhao-etal-2026-comprehensive}
Raoyuan Zhao, Yihong Liu, Hinrich Schuetze, and Michael~A. Hedderich. 2026.
\newblock \href {https://doi.org/10.18653/v1/2026.findings-eacl.276} {A comprehensive evaluation of multilingual chain-of-thought reasoning: Performance, consistency, and faithfulness across languages}.
\newblock In \emph{Findings of the {A}ssociation for {C}omputational {L}inguistics: {EACL} 2026}, pages 5223--5247, Rabat, Morocco. Association for Computational Linguistics.

\end{thebibliography}
\end{document}